%% file: main.tex
\documentclass[11pt, letterpaper]{cmu}

\usepackage[all]{hypcap}
\usepackage[comma,numbers,sort,compress]{natbib}
\usepackage{hyperref}[citecolor=magenta]
\usepackage{amsmath, amssymb}

\hypersetup{
    colorlinks = true,
    citecolor = {magenta},
}

\PassOptionsToPackage{numbers, compress}{natbib}

\usepackage[utf8]{inputenc}
\usepackage[T1]{fontenc}
\usepackage{url}
\usepackage{booktabs}
\usepackage{amsfonts}
\usepackage{nicefrac}
\usepackage{microtype}
\usepackage{xcolor}
\microtypecontext{spacing=nonfrench}
\usepackage{amsmath,amsthm}
\usepackage{graphicx}
\usepackage{mathtools}
\usepackage{mathrsfs}
\usepackage{dsfont}
\usepackage{float}
\usepackage{amssymb}
\usepackage{pifont}

\usepackage{xspace}
\usepackage[capitalize,noabbrev]{cleveref}
\usepackage{lipsum}
\usepackage{listings}
\usepackage{bbm}
\usepackage{tabularx}
\usepackage{lmodern}
\usepackage{multirow}
\usepackage[export]{adjustbox}
\usepackage{colortbl}
\usepackage{makecell}
\usepackage{wrapfig}
\usepackage{subcaption}

\newcolumntype{Y}{>{\raggedright\arraybackslash}X}

\usepackage[noend]{algpseudocode}
\newcommand{\algcommentcolor}{blue!70!black}
\algrenewcommand\algorithmiccomment[1]{\hfill{\color{\algcommentcolor}$\triangleright$~#1}}

\usepackage{setspace}
\usepackage{color}
\definecolor{deepblue}{rgb}{0,0,0.5}
\definecolor{deepred}{rgb}{0.6,0,0}
\definecolor{deepgreen}{rgb}{0,0.5,0}

\usepackage[most,skins,theorems,breakable]{tcolorbox}

\theoremstyle{plain}

\theoremstyle{definition}

\theoremstyle{remark}

\definecolor{rliableolive}{HTML}{BBCC33}
\definecolor{rliableblue}{HTML}{77AADD}
\definecolor{rliablered}{HTML}{EE8866}

\tcbset{
  aibox/.style={
    width=\linewidth,
    top=8pt,
    bottom=4pt,
    colback=rliableolive!8!white,
    colframe=black,
    colbacktitle=black,
    enhanced,
    center,
    attach boxed title to top left={yshift=-0.1in,xshift=0.15in},
    boxed title style={boxrule=0pt,colframe=white,},
  }
}
\newtcolorbox{AIbox}[2][]{aibox,title=#2,#1}

\usepackage{enumitem}

\newcommand\pythonstyle{\lstset{
basicstyle=\ttfamily\footnotesize,
language=Python,
morekeywords={self},
keywordstyle=\color{deepblue},
emph={MyClass,__init__},
emphstyle=\color{deepred},
stringstyle=\color{deepgreen},
frame=single,
showstringspaces=false
}}

\lstnewenvironment{python}[1][]
{
\pythonstyle
\lstset{#1}
}
{}

\newcommand\pythoninline[1]{{\pythonstyle\lstinline!#1!}}

\input{macro}

\makeatletter
\def\mathcolor#1#{\@mathcolor{#1}}
\def\@mathcolor#1#2#3{%
  \protect\leavevmode
  \begingroup
    \color#1{#2}#3%
  \endgroup
}
\makeatother

\usepackage[textsize=tiny]{todonotes}
\usepackage{algorithm}

\Crefformat{equation}{#2Eq.\;(#1)#3}
\Crefformat{figure}{#2Figure #1#3}
\Crefformat{assumption}{#2Assumption #1#3}
\Crefname{assumption}{Assumption}{Assumptions}

\usepackage{crossreftools}
\usepackage[suppress]{color-edits}

\usepackage{soul}
\definecolor{highlightmistake}{RGB}{255, 179, 179}
\definecolor{highlightcorrect}{RGB}{179, 255, 179}

\usepackage{bm}
\input{math_commands}

\usepackage{array}
\newcolumntype{C}[1]{>{\centering\arraybackslash}p{#1}}

\definecolor{myblue}{rgb}{0.1,0.4,0.9}

\definecolor{darkblue}{rgb}{0, 0, 0.5}
\definecolor{promptbg}{RGB}{232,242,255}
\definecolor{promptframe}{RGB}{92,140,214}
\definecolor{evalbg}{RGB}{241,252,241}
\definecolor{evalframe}{RGB}{77,162,94}
\definecolor{solutionbg}{RGB}{255,246,232}
\definecolor{solutionframe}{RGB}{214,139,42}
\definecolor{lightblue}{rgb}{0.22,0.45,0.70}
\newcommand{\parheading}[1]{\textbf{#1}}
\newcommand{\parhighlight}[1]{\textcolor{lightblue}{\textbf{\textit{#1}}}}

\newtcolorbox{promptbox}[1]{
  enhanced,
  breakable,
  arc=2.2mm,
  boxrule=0.8pt,
  left=1.3mm,right=1.3mm,top=1.0mm,bottom=1.1mm,
  colback=promptbg,
  colframe=promptframe,
  colbacktitle=promptframe!34,
  coltitle=black,
  fonttitle=\bfseries,
  title={#1}
}

\newtcolorbox{evalbox}[1]{
  enhanced,
  breakable,
  arc=2.2mm,
  boxrule=0.8pt,
  left=1.3mm,right=1.3mm,top=1.0mm,bottom=1.1mm,
  colback=evalbg,
  colframe=evalframe,
  colbacktitle=evalframe!30,
  coltitle=black,
  fonttitle=\bfseries,
  title={#1}
}

\newtcolorbox{smallevalbox}[1]{
  enhanced,
  breakable,
  arc=2.2mm,
  boxrule=0.8pt,
  left=1.3mm,right=1.3mm,top=1.0mm,bottom=1.1mm,
  colback=evalbg,
  colframe=evalframe,
  colbacktitle=evalframe!30,
  coltitle=black,
  fonttitle=\bfseries,
  title={#1}
}

\newtcolorbox{solutionbox}[1]{
  enhanced,
  breakable,
  arc=2.2mm,
  boxrule=0.8pt,
  left=1.3mm,right=1.3mm,top=1.0mm,bottom=1.1mm,
  colback=solutionbg,
  colframe=solutionframe,
  colbacktitle=solutionframe!34,
  coltitle=black,
  fonttitle=\bfseries,
  title={#1}
}

\usepackage{textcomp}
\usepackage{upquote}
\lstdefinestyle{mypython}{
  language=Python,
  basicstyle=\ttfamily\small,
  keywordstyle=\color{blue!70!black}\bfseries,
  stringstyle=\color{green!45!black},
  commentstyle=\color{gray}\itshape,
  showstringspaces=false,
  breaklines=true,
  breakatwhitespace=true,
  columns=fullflexible,
  frame=single,
  framesep=4pt,
  xleftmargin=10pt,
  literate={`}{{\textasciigrave}}1
}

\title{\methodname: Training Robots to Reason in Natural Language via Reinforcement Learning}

\author[1]{Lehong Wu}
\author[1]{Yuxiao Qu}
\author[1]{Zheyuan Hu}
\author[1]{Ivan Zhang}
\author[1]{Limin Wei}
\author[1]{Zackory Erickson}
\author[1]{Aviral Kumar}
\affil[1]{Carnegie Mellon University}

\correspondingauthor{lehongw2@andrew.cmu.edu}

\input{sections/abstract}

\begin{document}

\maketitle
\vspace{-0.3cm}

\input{sections/introduction}

\input{sections/related_work}
\input{sections/recipe}

\input{sections/exp}

\input{sections/conclusion}

\input{sections/acknowledgment}
\bibliography{main}

\appendix
\onecolumn
\part*{Appendices}

\input{appendix/exp_details}
\input{appendix/additional_exp}

\input{appendix/future_work}
\input{appendix/examples}
\input{appendix/prompts}

\end{document}

%% file: macro.tex
\newcommand{\bz}{\mathbf{z}}

\definecolor{blanchedalmond}{rgb}{1.0, 0.92, 0.8}
\definecolor{carmine}{rgb}{0.59, 0.0, 0.09}
\definecolor{lightblue}{rgb}{0.22,0.45,0.70}

\renewcommand{\mathbf}{\boldsymbol}

\makeatletter
\def\Ddots{\mathinner{\mkern1mu\raise\p@
\vbox{\kern7\p@\hbox{.}}\mkern2mu
\raise4\p@\hbox{.}\mkern2mu\raise7\p@\hbox{.}\mkern1mu}}
\makeatother

\numberwithin{equation}{section}

\definecolor{amaranth}{rgb}{0.9, 0.17, 0.31}
\definecolor{antiquebrass}{rgb}{0.8, 0.58, 0.46}
\definecolor{antiquefuchsia}{rgb}{0.57, 0.36, 0.51}
\definecolor{chromeyellow}{rgb}{0.31, 0.47, 0.26}

\usepackage[dvipsnames]{xcolor}
\definecolor{maj5}{HTML}{2b8cbe}
\definecolor{maj5Imp}{HTML}{084081}
\definecolor{seq5wo}{HTML}{d95f0e}
\definecolor{seq5woImp}{HTML}{662506}
\definecolor{seq5w}{HTML}{6a51a3}
\definecolor{seq5wImp}{HTML}{3f007d}
\definecolor{selfwo}{HTML}{d95f0e}
\definecolor{selfwoImp}{HTML}{662506}
\definecolor{selfw}{HTML}{6a51a3}
\definecolor{selfwImp}{HTML}{3f007d}
\definecolor{glorewo}{HTML}{d95f0e}
\definecolor{glorewoImp}{HTML}{662506}
\definecolor{glorew}{HTML}{6a51a3}
\definecolor{glorewImp}{HTML}{3f007d}
\definecolor{vstar}{HTML}{d95f0e}
\definecolor{vstarImp}{HTML}{662506}

%% file: math_commands.tex
\usepackage{amsmath,amsfonts,bm}

\def\eqref#1{Eq.~\ref{#1}}

\def\1{\bm{1}}

\DeclareMathAlphabet{\mathsfit}{\encodingdefault}{\sfdefault}{m}{sl}
\SetMathAlphabet{\mathsfit}{bold}{\encodingdefault}{\sfdefault}{bx}{n}

\newcommand{\methodname}{$\mathcal{R}^3$}

\renewcommand{\bz}{\mathbf{z}}

\newcommand{\by}{\mathbf{y}}

\newcommand{\bx}{\mathbf{x}}

\newcommand{\taskSetMidTrain}{\ensuremath{\mathcal{T}_{\mathrm{M}}}}
\newcommand{\taskSetPostTrain}{\ensuremath{\mathcal{T}_{\mathrm{R}}}}
\newcommand{\taskSetOOD}{\ensuremath{\mathcal{T}_{\mathrm{O}}}}

\newcommand{\taskgroup}{\texttt{group}}
\newcommand{\taskline}{\texttt{line}}
\newcommand{\taskV}{\texttt{V}}
\newcommand{\taskL}{\texttt{L}}
\newcommand{\taskclearqtr}{\texttt{clear\_qtr}}
\newcommand{\taskiip}{\texttt{iip}}

\newcommand{\taskT}{\texttt{T}}
\newcommand{\taskgris}{\texttt{gris}}
\newcommand{\taskiV}{\texttt{iV}}

\newcommand{\taskdiagline}{\texttt{diag\_line}}
\newcommand{\taskrect}{\texttt{rect}}
\newcommand{\taskmid}{\texttt{mid}}
\newcommand{\taskiL}{\texttt{iL}}
\newcommand{\taskclearhalf}{\texttt{clear\_half}}
\newcommand{\taskspack}[1]{\texttt{Task #1}}

\newcommand{\ours}{\methodname}
\newcommand{\oursMidOnly}{\methodname~(mid only)}
\newcommand{\oursRLOnly}{\methodname~(RL only)}
\newcommand{\oursQuarterMid}{\methodname~(1/4th mid)}
\newcommand{\ILMid}{IL~(mid only)}
\newcommand{\IL}{IL}

\newcommand{\ILPretrain}{IL~(Pre-train)}
\newcommand{\ILCotrain}{IL~(Co-train)}

%% file: sections/abstract.tex
\begin{abstract}
\textbf{Abstract:} 
Reasoning in language allows foundation models to spend more test-time compute on hard problems, such as those requiring decomposition, constraint tracking, and prediction of future consequences. Whether this mechanism can improve robotic manipulation remains unclear, where long-horizon tasks require tracking partial progress, reasoning about object relations, recovering from mistakes, and steering noisy low-level policies. In this paper, we study whether VLMs can be trained to reason directly in natural language to guide low-level manipulation policies. 
We introduce \methodname, a simple post-training recipe that turns off-the-shelf VLMs into robotic reasoners: it first mid-trains a VLM on expert-generated reasoning traces to initialize the desired reasoning style, then improves the reasoner with single-step rubric-based RL from offline action data. 
Unlike prior robotic reasoning methods that mostly use structured traces as auxiliary supervision, \methodname{} trains free-form language reasoning to produce test-time guidance for action. 
We instantiate \methodname{} on Language Table and simulated bimanual grocery packing, two controlled testbeds for studying robotic reasoning and long-horizon manipulation.
\methodname{} improves exploration and generalization across unseen tasks and significantly outperforms instruction-only imitation learning baselines on both benchmarks.
Our analyses suggest that free-form language reasoning can function as a test-time compute mechanism for steering low-level policies.
Our project page is available at
\href{https://robotic-reasoner.github.io/}{https://robotic-reasoner.github.io/}.
\end{abstract}

%% file: sections/introduction.tex
\section{Introduction}
\label{sec:introduction}

Reasoning in natural language provides an effective mechanism for spending more compute on harder test problems, and offers a data-efficient recipe for broad generalization. This recipe is clearly useful in many domains including visual perception~\citep{fan2026stepcotstepwisevisualchainofthought,chen2023seethinkconfirminteractive}. More broadly, even when the final output is not language~\citep{driess2023palmeembodiedmultimodallanguage}, language reasoning can help a model decompose the problem, identify relevant constraints, and make reliable predictions. Robotic manipulation is therefore a natural domain for reasoning based foundation models: manipulation requires interpreting a scene, understanding physical constraints, anticipating the effect of action on future parts of the trajectory, and acting conditioned on this understanding. One might expect that training robotic policies to reason before acting would improve generalization by allowing the model to spend test-time compute on the problem instance.

Reasoning for robotic manipulation has already received a fair bit of attention. Recent generalist robot policies and vision-language-action models, including ECoT~\citep{zawalski2025robotic}, SteerVLA~\citep{gao2026steervlasteeringvisionlanguageactionmodels}, and MolmoAct~\citep{lee2025molmoact}, incorporate intermediate representations ranging from object-centric annotations and short plans to depth-aware perception tokens and image-space trajectories. Complementary approaches expose generalist policies through semantic interfaces: $\pi_{0.7}$ can be steered at inference time with subtask instructions and visual subgoals~\citep{intelligence2026pi07steerablegeneralistrobotic}, while SARL uses online RL to learn a high-level policy over language commands that steer a fixed VLA through long-horizon tasks~\citep{bhatia2026adaptinggeneralistrobotpolicies}. However, these approaches do not train free-form natural-language reasoning of the kind that has proven effective in language. Prior work shows that structured reasoning supervision can improve grounding and perception~\citep{chen2025training},
but these gains appear to arise mainly from training-time supervision: after training with reasoning, generating reasoning at test time provides little additional benefit~\citep{chen2025training,fang2026molmoact2}. Thus, existing work establishes structured CoT as an auxiliary training signal, but leaves open whether flexible language reasoning can serve as a mechanism for spending test-time compute for manipulation, and how to learn to do that in reality.

\begin{figure}[t]
\centering
\vspace{-0.2cm}
\includegraphics[
  width=\linewidth,
  trim=0cm 27cm 5.75cm 0cm,
  clip
]{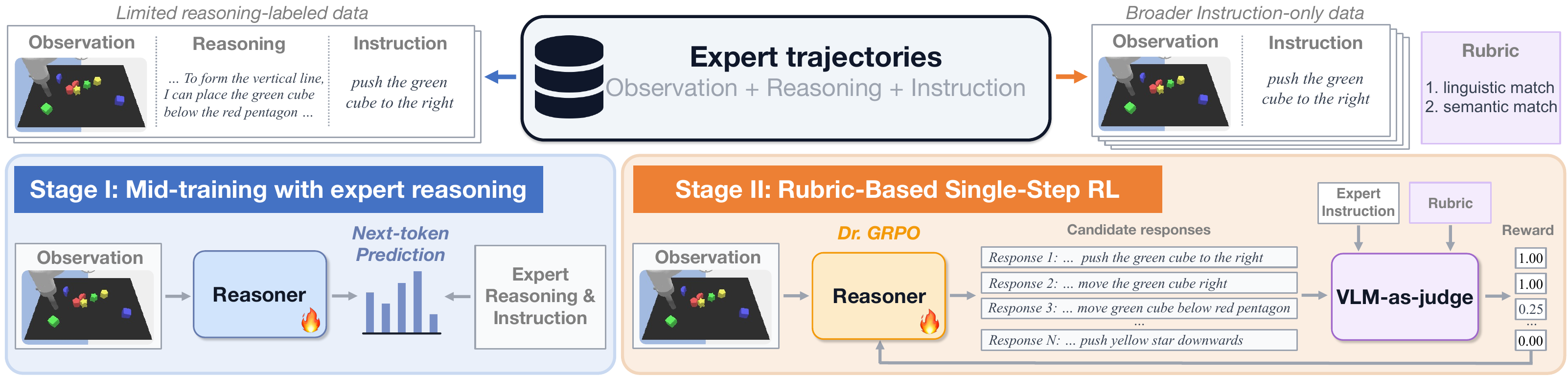}
\vspace{-0.3cm}
\caption{\footnotesize{\textbf{Two-stage training of \methodname.} \methodname{} trains a high-level VLM to reason in natural language and steer a fixed low-level robot policy for robotic manipulation tasks. \methodname{} proceeds in two stages: Stage I (mid-training) imbues an off-the-shelf VLM with the reasoning style and behaviors needed to produce useful instructions for the low-level policy. Stage II (single-step RL) further improves the VLM by training it to generate reasoning traces that match the expert instruction in offline data.}}\label{fig:workflow}
\vspace{-0.3cm}
\end{figure}

We study how to train vision-language models (VLMs) to use free-form natural language as a mechanism to spend test-time compute for robotic manipulation. Our main idea is to turn expert data into supervision for VLM reasoning. Given a scene, interaction history, and an expert action, we post-train a VLM to produce language-based reasoning that allows it to arrive at a semantically similar instruction to the expert’s. A language-conditioned low-level policy then takes this VLM's instruction and outputs the action that directly controls the robot.
This training procedure gives the VLM reasoning capabilities that allow it, at test time, to produce language guidance for steering the low-level policy. To instantiate this idea and study key design choices behind it, we focus on the \emph{Language Table}~\citep{lynch2022interactivelanguagetalkingrobots}  environment and a long-horizon grocery packing environment~\citep{anonymous2026vlaexplore}, which provide controlled settings for studying visual reasoning, language-conditioned manipulation, and long-horizon planning, along with a pretrained steerable policy that reliably follows a wide range of instructions. 

Our approach, \methodname, uses a two-stage recipe inspired by LLM post-training, as shown in Figure~\ref{fig:workflow}. We first generate multi-turn manipulation trajectories by prompting expert reasoners to steer the low-level policy while recording their reasoning. These trajectories contain partial progress, mistakes, recoveries, and alternative action choices, producing contexts where the model must reason over past interaction rather than only the current frame. Stage I \emph{mid-trains} a VLM on expert-generated reasoning traces, regardless of trajectory success, to initialize the reasoning style needed for manipulation. Stage II then improves this reasoner with \emph{single-step reinforcement learning (RL) from offline data}. Here, we no longer assume access to expert reasoning traces: conditioned on the scene and interaction history, the model generates reasoning and an instruction, and is rewarded by a rubric-based VLM judge~\citep{dong2026musebenchmarkingmanufacturablefunctional} when its instruction semantically matches the expert's. This yields a practical recipe: use limited reasoning-labeled trajectories to initialize the reasoner, then use more instruction-only offline data to further improve it.

Beyond the method itself, this framework lets us study key design choices for robotic reasoning: how to collect demonstrations, condition on history, initialize the reasoner, and which RL formulation can improve reasoning from action supervision.
Empirically, we find that \methodname~improves high-level steering across both seen and unseen Language Table tasks, outperforming instruction-only imitation learning.
Similarly, on the grocery packing environment, \oursRLOnly{} outperforms instruction-only imitation without reasoning. Our analyses show that \methodname~learns more deliberate action-oriented reasoning behaviors: it compares alternative choices, re-examines the scene and history when uncertain, and selects next steps more incrementally.
Through VQA diagnostics, comparisons with non-reasoning policies with auxiliary reasoning supervision, and controlled interventions on the reasoning budget, we find that explicit inference-time reasoning improves generalization beyond using reasoning only as a training-time supervision signal. %
Together, these results provide evidence that language reasoning causally contributes to task performance and serves as useful test-time compute.

%% file: sections/related_work.tex
\section{Related Work}
\label{sec:related_work}

\parheading{Reasoning and test-time compute.}
Chain-of-thought reasoning and test-time scaling methods show that language models can solve harder problems by spending extra compute on reasoning before answering~\citep{wei2023chainofthoughtpromptingelicitsreasoning,kojima2023largelanguagemodelszeroshot,wang2023selfconsistencyimproveschainthought,yao2023treethoughtsdeliberateproblem,snell2024scalingllmtesttimecompute,yang2025thinkingoptimalscalingtesttimecompute}. Similar ideas extend to VLMs, where textual rationales, grounded explanations, and visual chain-of-thought traces improve VQA and multimodal reasoning~\citep{lu2022learnexplainmultimodalreasoning,zhang2024multimodal,shao2024visual,zhang2025improve}. However, these works mainly evaluate static question answering rather than long-horizon interaction: reasoning about a static image need not transfer to embodied action. Indeed, our experiments show that VLMs with similar static VQA performance can differ substantially in steering a low-level policy on long-horizon manipulation tasks that require reasoning. Reasoning has also helped in interactive domains such as coding and web agents~\citep{wei2025webagentr1trainingwebagents,cho2025selfcorrectingcodegenerationusing}, where models call tools, observe feedback, and revise their behavior online~\citep{schick2023toolformerlanguagemodelsteach,ma2026act2seeemergentactivevisual,shinn2023reflexionlanguageagentsverbal}. Robotic manipulation differs because reasoning must be grounded in a low-level controller that often induces substantial partial observability for the high-level reasoner.

\parheading{Intermediate structures in robotic manipulation.}
Robotics has long used intermediate structure between perception and action. Task-and-motion planning combines symbolic task reasoning with geometric motion planning~\citep{kaelbling2013integrated,garrett2020integratedtaskmotionplanning,guo2023recent}; visual foresight predicts future observations for planning~\citep{finn2017deep,ebert2018visual}; and hierarchical or latent-planning methods learn reusable skills from demos or play~\citep{lynch2020learning,rosete2023latent}. Language-based systems such as SayCan~\citep{ahn2022can}, Inner Monologue~\citep{huang2022inner}, and Code as Policies~\citep{liang2023code} use LLMs for decomposition, feedback, or program synthesis, while relying on external skills, values, or controllers. These works show the value of reasoning before acting, but the reasoning is typically symbolic, predictive, latent, modular, or hand-designed. In contrast, we train a VLM itself to produce natural-language reasoning that steers a frozen low-level language-conditioned policy at test time.

\parheading{Reasoning in generalist robot policies.}
Recent generalist robot policies and vision-language-action models, including RT-2, Octo, OpenVLA, GR00T, GR-3, and Gemini Robotics, use VLM backbones pretrained on internet data but do not incorporate explicit reasoning~\citep{brohan2023rt2visionlanguageactionmodelstransfer,team2024octo,kim2024openvla,bjorck2025gr00t,cheang2025gr3technicalreport,geminiroboticsteam2025geminiroboticsbringingai}. Some subsequent works use reasoning-like signals only as training-time supervision, such as grounded reasoning traces, object detections, or semantic subtask predictions~\citep{chen2025training,intelligence2025pi05visionlanguageactionmodelopenworld}.
Others additionally produce or consume inference-time intermediates, such as language guidance, plans, subtasks, constraints, affordances, history summaries, visual traces, depth representations, or subgoal images before action generation~\citep{clark2025actionfreereasoningpolicygeneralization,niu2025llarva,huang2025rekep,zheng2025tracevla,dai2024racerrichlanguageguidedfailure,zawalski2025robotic,yang2025instructvla,lin2025onetwovla,shi2025hirobotopenendedinstruction,intelligence2026pi07steerablegeneralistrobotic,lee2025molmoact,fang2026molmoact2,gao2026steervlasteeringvisionlanguageactionmodels,li2025coavla,zhao2025cotvla,li2025hamster,bhatia2026adaptinggeneralistrobotpolicies}.
In contrast, we isolate reasoning as a training-design problem: we train a VLM reasoner to generate free-form natural-language reasoning as guidance, rather than relying on hand-designed intermediate representations, while keeping the low-level policy frozen.

%% file: sections/recipe.tex
\vspace{-0.25cm}
\section{\methodname:   \underline{R}obotic \underline{R}easoners via \underline{R}einforcement Learning}
\label{sec:recipe}
\begin{wrapfigure}{r}{0.35\textwidth}
\centering
\vspace{-0.1cm}
\includegraphics[
  width=\linewidth,
  trim=0cm 15.5cm 40cm 0cm,
  clip
]{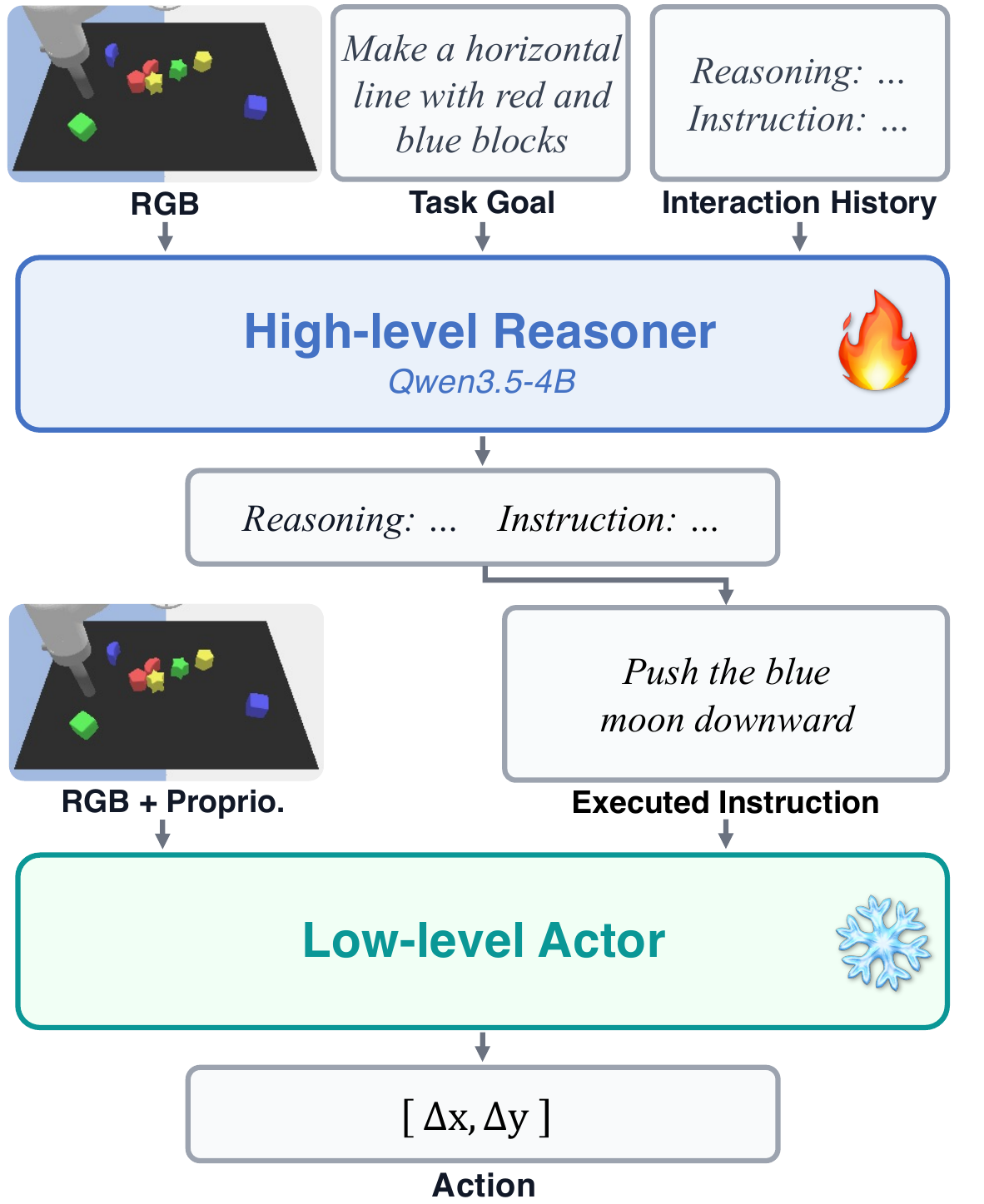}
\vspace{-0.7cm}
\caption{\footnotesize{\textbf{Policy architecture.} A high-level VLM generates a reasoning trace and an instruction given the scene, goal, and previous response. A language-conditioned low-level actor takes the instruction as input and emits the action that controls the robot.}}
\label{fig:architecture}
\vspace{-0.35cm}
\end{wrapfigure}

\vspace{-0.2cm}
To train reasoners for robotic manipulation, we use a hierarchical architecture as in prior work~\citep{shi2025hirobotopenendedinstruction}: a low-level policy controls the robot, while a high-level VLM provides instructions that steer this policy. Figure~\ref{fig:architecture} shows our specific architecture. Our goal is to train the high-level VLM to reason in natural language before issuing an instruction. 
We describe our problem setup first and then our approach.

\vspace{-0.1cm}
\subsection{Problem Setup and Environments}
\vspace{-0.1cm}
We model each task as a decision process \(\mathcal{M}_g = (\mathcal{S}, \mathcal{A}, P, r_g)\), where \(g\) is a textual long-horizon goal, \(s_t \in \mathcal{S}\) contains visual and proprioceptive information, \(a_t \in \mathcal{A}\) is a robot action, \(P\) is the transition dynamics, and \(r_g\) is a binary success reward. For example, \(g\) might be ``make a V-shape with the red moon, blue cube, and yellow star.'' We assume access to a pretrained low-level policy \(\pi_{\mathrm{lo}}(a_t | s_t, u_t)\), where \(u_t \in \mathcal{U}\) is a \emph{short-horizon} subtask instruction such as ``move the red block left'' or ``push the blue cube to the green star''. We train a high-level VLM \(\pi_\theta(\bz_t, u_t | \bx_t, g)\), where \(\bx_t\) is the history context, \(\bz_t\) is the reasoning trace, and \(u_t\) is the instruction to the low-level policy. At each step, the high-level VLM uses \(\bz_t\) to reason about the progress of the task and the consequences of the action, emits \(u_t\), and the low-level policy executes a fixed-length action chunk \(a_t\), with \(\bz_t, u_t \sim \pi_\theta(\cdot | \bx_t, g)\) and \(a_t \sim \pi_{\mathrm{lo}}(\cdot | s_t, u_t)\).

Training the reasoner requires design choices that are central to any learning-based robotic system: \textbf{(1)} what data should be used, and how the reasoning process should be parameterized; \textbf{(2)} how to initialize or warm-start the reasoner; \textbf{(3)} what objective can improve reasoning beyond simply memorizing reasoning from experts. In this section, we develop \methodname, a recipe for training robotic reasoners. At a high level, \methodname{} collects high-coverage expert trajectories on diverse tasks, mid-trains the VLM on a small portion of expert reasoning traces to initialize the desired reasoning style, and further improves it with rubric-based RL from more offline non-reasoning expert data on a broader set of tasks. This turns expert manipulation demonstrations into practical training signals for reasoning, while avoiding expensive robot rollouts.

\parheading{Data collection for Language Table.}
Training a reasoner for manipulation requires more than states, actions, and subtask annotations. Standard teleoperated demonstrations, even when postprocessed with subtasks or instructions~\citep{intelligence2025pi05visionlanguageactionmodelopenworld}, do not explain why an expert chooses the subtask, how it interprets partial progress, or how it recovers from mistakes. We need data that covers diverse behaviors (including behaviors that showcase recovery and imperfect attempts at manipulation or high-level planning) and intermediate states, with reasoning traces that explain high-level decisions. The tasks themselves must also require heavy reasoning: if they are too short-horizon or solvable from the current frame, reasoning provides little benefit and may be discarded during post-training~\citep{feng2026procvlmlearningproceduregroundedprogress}. Unfortunately, a number of simulated environments test short-horizon performance in relatively simple scenes, with no clear room for benefiting from reasoning in language. In contrast, Language Table provides cluttered scenes with somewhat imperfect low-level policies, and this requires a reasoning system to deliberately evaluate multiple courses of action to succeed at the task. 

We therefore design 14 types of long-horizon block arrangement tasks in Language Table that require composing object movements and reasoning about spatial relationships. The task suite is designed to test relational transfer, compositional generalization, and present increasing geometric difficulty. 
Such training data could be collected from human experts verbalizing their thought process while performing teleoperation, but this is expensive and difficult to scale for statistically significant results in a controlled study\footnote{That said, a protocol that records the ``stream of consciousness'' of a human teleoperator using a microphone and rewrites this data with off-the-shelf VLMs can be utilized for seeding reasoning behavior.}. 
Thus, we emulate this setting with an expert VLM reasoner: given a task goal, the current observation, and the interaction history, the VLM produces a reasoning trace followed by a short-horizon instruction for steering the low-level policy. The low-level policy executes the action, the environment transitions, and the process repeats.

We use Gemini 3 Flash as the ``human'' expert and construct two data subsets based on the supervision exposed to the learner. The first subset exposes expert reasoning traces together with the instructions, and is used to seed reasoning behaviors during \emph{mid-training} (Section~\ref{subsec:stage1}). The second subset exposes only expert instructions, with reasoning traces withheld, and is used for \emph{RL} (Section~\ref{subsec:stage2}). This emulates the supervision regime we target: high-quality reasoning traces are expensive, while subtask-level instruction labels are easier to obtain. In RL, the model must generate its own reasoning and receives reward by comparing its predicted instruction against the expert's. This separation also aligns with practical constraints on data collection: while expert reasoning is difficult to obtain for all transitions, expert instructions are readily available in many settings.

\begin{figure}[t]
\centering
\includegraphics[width=\linewidth, 
trim=0 0cm 0cm 0,
clip
]{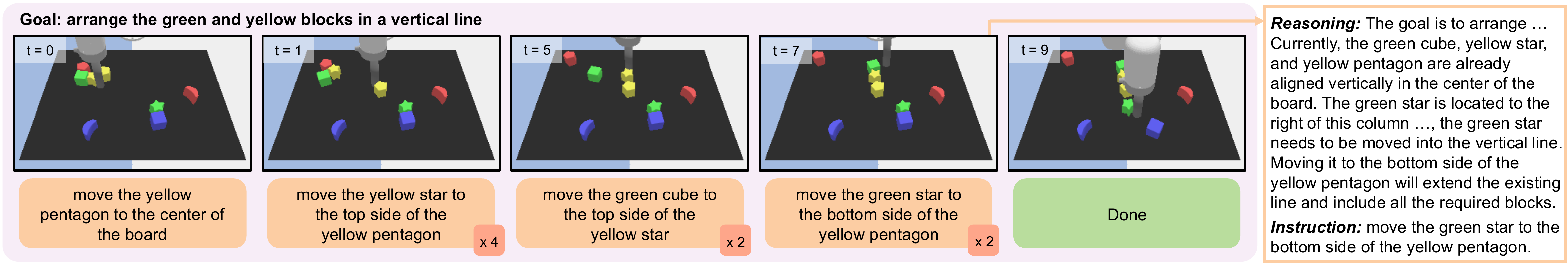}
\caption{\footnotesize{\textbf{Example of expert-collected trajectory and reasoning trace on the \taskline~task of Language Table.} The notation \(\times n\) indicates that the expert repeats this instruction $n$ times.}}
\label{fig:expert_example}
\end{figure}

\begin{wrapfigure}{r}{0.45\textwidth}
\centering
\vspace{-0.35cm}
\includegraphics[width=\linewidth]{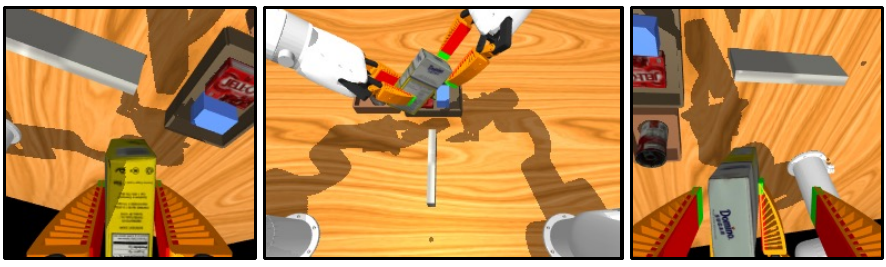}
\vspace{-0.55cm}
\caption{\footnotesize{\textbf{Example grocery packing task.} From left to right: left-wrist, base, and right-wrist camera views.}}
\label{fig:packing_views}
\vspace{-0.15cm}
\end{wrapfigure}
\parheading{Data collection for grocery packing.} To demonstrate the generality of our findings, we also apply our approach to a bimanual long-horizon grocery packing task suite \citep{anonymous2026vlaexplore}. These data are collected by human operators via teleoperation across a set of simulated task goals (see Figure~\ref{fig:packing_views} for examples), rather than by a Gemini expert. The collected data are then post-processed into segments, each labeled with a short-horizon instruction that specifies the task performed in that segment. In this setup, we do not assume access to reasoning traces and rely on direct RL on top of the base VLM (i.e., ``RL zero''~\citep{deepseekr1}) to improve performance. Thus, this setting removes the need to collect reasoning traces altogether.

\begin{wraptable}{r}{0.35\linewidth}
\centering
\small
\setlength{\tabcolsep}{6pt}
\renewcommand{\arraystretch}{1.05}
\begin{tabular}{@{}l c c@{}}
\toprule
\textbf{Task} & \textbf{w/o history} & \textbf{w/ history} \\
\midrule
\taskline & 44.9\% & 51.0\% \\
\taskV    & 52.3\% & 57.6\% \\
\bottomrule
\end{tabular}
\caption{\footnotesize{\textbf{Ablation of interaction history on the expert.} Incorporating history improves the expert VLM's performance.}}
\label{tab:gemini_history_ablation}
\vspace{-0.7cm}
\end{wraptable}
\parheading{Input to the VLM.} Humans naturally rely on memory when solving long-horizon (manipulation) tasks, often maintaining a coherent plan and reusing or refining previous actions. Without history context, the expert lacks these behaviors. We therefore collect data both with and without history. History can be represented in several ways, including past frames, past responses, and learned summaries. In this work, we instantiate history as the full response from the previous step. This allows the VLM to carry forward its inferred progress and plan. We evaluate Gemini 3 Flash, the expert reasoner, on two representative tasks, \taskline{} and \taskV~(see Appendix~\ref{app:exp_details} for task descriptions), with and without history in the context. As shown in Table~\ref{tab:gemini_history_ablation}, history context consistently improves pass@1: from 44.9\% to 51.0\% on \taskline, and from 52.3\% to 57.6\% on \taskV. These gains suggest that history helps to track progress, resolve ambiguities, and preserve a coherent plan. Therefore, we use history for our data collection. Examples of the collected trajectories are shown in Figure~\ref{fig:expert_example} and Appendix~\ref{app:examples}.

\subsection{Stage I: Mid-Training Reasoning Behaviors into the VLM}
\label{subsec:stage1}
In preliminary experiments, we found that even the strongest open-source VLMs at model sizes suitable for real-time robotic control, i.e., under 10B parameters, did not naturally produce the style of reasoning needed for our manipulation tasks.
Their reasoning was often shallow: it mentioned generic steps such as identifying objects or moving toward the goal, but failed to track task progress, object relations, failed attempts, or constraints that are critical for selecting the next subtask. In Appendix~\ref{app:examples}, we observe that the base model either fails to use spatial relationships between blocks or blindly trusts that the previous instruction was executed correctly, and these behaviors inhibit it from solving the task. We also observed substantial thought-switching, a failure mode related to ``underthinking''~\citep{wang2025thoughtsplaceunderthinkingo1like}. These observations highlight the need for a warm-up training stage before off-the-shelf VLMs can serve as reliable robotic reasoners.

Pretrained models often cannot produce useful or productive reasoning even when they can generate natural-sounding chains of thought. A common fix is \emph{mid-training}: a phase that does not optimize reward directly, but instead initializes the model with useful reasoning patterns or priors such as decomposition, constraint tracking, and self-correction~\citep{qu2024recursiveintrospectionteachinglanguage}.
We adopt the same idea for robotic reasoning by mid-training the VLM on expert-generated reasoning traces. This mirrors the role of mid-training in language and math reasoning, where the goal is to expose the model to useful reasoning patterns before RL~\citep{qu2025optimizingtesttimecomputemeta,wang2025octothinkermidtrainingincentivizesreinforcement}.
Concretely, each mid-training example consists of an interaction context \(\bx_t\), an expert reasoning trace \(\bz_t\), and a high-level instruction \(u_t\). Let \(\by_t = (\bz_t, u_t)\) denote the target sequence formed by concatenating the reasoning and instruction. We train with a standard next-token prediction objective:
\begin{align}
\mathcal{L}_{\mathrm{SFT}}(\theta)
=
-\mathbb{E}_{(\bx_t,\by_t)\sim\mathcal{D}_{\mathrm{SFT}},\; i\sim \mathrm{Unif}(\{1,\ldots,|\by_t|\})}
\left[
\log p_\theta\!\left(\by_{t,i}\mid \bx_t, \by_{t,<i}\right)
\right].
\end{align}
This teaches the model to reason about the state and interaction history before emitting the instruction for the low-level policy. We train on both successful and unsuccessful reasoning traces. Successful trajectories show how reasoning leads to useful instructions, while unsuccessful trajectories still provide supervision about partial progress, mistakes, and recovery attempts.

\subsection{Stage II: Rubric-Based Single-Step RL with Offline Data}
\label{subsec:stage2}
While mid-training seeds the VLM with the desired reasoning behaviors, it does not ensure that test-time reasoning produces effective instructions for steering the low-level policy. A natural next step to optimize for reasoning behavior would be online RL: run \emph{multi-turn} rollouts that interleave calls to the high-level VLM reasoner with executions of the low-level policy, and use final task success as the reward. However, this requires expensive environment interaction and long-horizon credit assignment. Credit assignment is especially difficult in our hierarchical setting, where failures can arise from poor reasoning, ambiguous high-level instructions, or bad instruction following of the fixed low-level policy.

We therefore use a single-step formulation for RL on an expert dataset consisting of an interaction history \(\bx_t\) and the corresponding expert instruction \(u_t^\star\), with no expert reasoning. The model samples \((\bz_t,u_t) \sim \pi_\theta(\cdot|\bx_t,g)\), where \(\bz_t\) is a free-form reasoning and \(u_t\) is the low-level instruction. We reward the model when \(u_t\) is semantically consistent with \(u_t^\star\), and when \(\bz_t\) explains why this instruction is appropriate given the scene and interaction history. Thus, single-step RL improves the reasoner from action supervision, without requiring human-written rewards or multi-turn robot rollouts.

The reward function receives \(g\), \(\bx_t\), \(u_t^\star\), and \((\bz_t,u_t)\), and returns a scalar reward
\(R(\bx_t,g,u_t^\star,\bz_t,u_t)\). 
It can be instantiated either as a VLM judge guided by rubrics or a verifiable reward. While a verifiable reward as simple as string matching is easy to implement, VLM-as-a-judge allows for more flexible answer formats and more detailed scoring criteria.
We also assign a negative reward to overly short responses to avoid degenerate traces that skip reasoning and jump directly to the final instruction. 
We then train the VLM reasoner with Dr.GRPO~\citep{liu2025understandingr1zeroliketrainingcritical}. For each context \(\bx_t\), we sample a group of \(K\) responses \(\{(\bz_t^{(k)},u_t^{(k)})\}_{k=1}^K\), score them each with the reward \(R^{(k)}\).
We then optimize a policy gradient loss:
\begin{align}
\mathcal{L}_{\mathrm{GRPO}}(\theta)
=
-\mathbb{E}_{t,k}
\left[
\min\left(
\rho_t^{(k)} A^{(k)},
\mathrm{clip}(\rho_t^{(k)}, 1-\epsilon_{\mathrm{clip}}, 1+\epsilon_{\mathrm{clip}}) A^{(k)}
\right)
\right],
\end{align}
where \(\rho_t^{(k)}
=
\frac{
\pi_\theta(\bz_t^{(k)},u_t^{(k)}|\bx_t,g)
}{
\pi_{\theta_{\mathrm{old}}}(\bz_t^{(k)},u_t^{(k)}|\bx_t,g)
}\) denotes the importance-sampling ratio between the current and old policy, and \(A^{(k)} = R^{(k)} - \frac{1}{K}\sum_{j=1}^K R^{(j)}\) is the advantage obtained by normalizing rewards within the group.

\begin{AIbox}{Summary: Training Robotic Reasoners via Reinforcement Learning (\methodname{})}
\begin{itemize}[leftmargin=*, itemsep=1pt, topsep=2pt]
  \item \methodname{} trains a high-level VLM to reason before instructing a pretrained low-level robot policy.
  \item We mid-train on a small reasoning-labeled subset to initialize useful reasoning behaviors.
  \item We then apply single-step RL on broader offline data containing only expert instructions.
\end{itemize}
\end{AIbox}

\parheading{RL design choices for Language Table.} We use VLM-as-a-judge as the reward function on Language Table because the valid instruction set is not finite, and multiple instructions can be equivalent to steering the policy.
The VLM judge, Qwen3.5-35B-A3B, follows rubrics that evaluate whether the predicted instruction aligns with the expert's intent, is feasible for the low-level policy, and would lead to a similar outcome. Therefore, the reward reflects semantic matching rather than string matching. We provide the detailed reward in Appendix~\ref{app:subsec:reward} and the rubrics in Appendix~\ref{app:subsec:vlm_as_judge_prompt}. 
We validate the VLM judge against human labels in Appendix~\ref{app:judge_validation}.
On 100 prompt--response pairs scored by three human annotators, our judge agrees closely with the human-majority labels, approaching inter-annotator agreement. Alternative VLM judges perform similarly, indicating that our reward is reliable and insensitive to judge choice.

In particular, we highlight two other design choices: \textbf{(1)} \emph{Reasoning context imputation.} The RL data contains expert instructions but not expert reasoning traces. Because the previous response is part of the interaction history, we impute missing reasoning by sampling 48 responses from the mid-trained model at each step. If any sampled response results in the same instruction as the expert at the previous state, we use it as the previous-step context; otherwise, we provide only the previous instruction with no reasoning. \textbf{(2)} \emph{Filtering repetitive steps.} In our preliminary RL experiments, the VLM reasoner often learned to repeat the previous instruction, since many expert trajectories contain repeated instructions, and repetition can become a severe reward shortcut during RL. We therefore remove repetitive steps from RL data so that training focuses on meaningful instruction changes. We observe that the fraction of repeated instructions of the VLM decreases only slightly after RL, suggesting that this data filtering does not prevent the model from repeating instructions when appropriate.

\parheading{RL design choices for grocery packing.} 
Packing instructions form a finite set of pack / remove / transfer commands, each without semantic ambiguity, so the policy can be prompted and trained to match the exact ground-truth instruction. We therefore use a simple exact-match reward instead of VLM-as-a-judge: \(1.0\) if the parsed instruction string equals the expert instruction, and \(0.0\) otherwise.
We also empirically found that instantiating interaction history as the previous response or the previous instruction yields comparable performance, so we simply use the previous instruction, thereby eliminating the need for the reasoning context imputation process before RL.

%% file: sections/exp.tex
\vspace{-0.2cm}
\section{Experimental Evaluation on Language Table}
\label{sec:result}

\newcommand{\best}[1]{\textbf{#1}}
\newcommand{\second}[1]{\underline{#1}}
\newcommand{\ind}[1]{\hspace{0.7em}#1}
\newcommand{\smallci}[1]{\scalebox{0.62}{$\pm$\,#1}}
\newcommand{\valci}[2]{#1\,\smallci{#2}}
\newcommand{\tabdlt}[1]{\scalebox{0.62}{$(#1)$}}
\definecolor{deltagreenlight}{RGB}{238,249,241}
\definecolor{deltagreen}{RGB}{220,241,227}
\definecolor{deltared}{RGB}{248,227,227}
\newcommand{\deltapos}[2]{\cellcolor{deltagreenlight}\valci{$+$#1}{#2}}
\newcommand{\deltaposstrong}[2]{\cellcolor{deltagreen}\valci{$+$#1}{#2}}
\newcommand{\deltanegstrong}[2]{\cellcolor{deltared}\valci{$-$#1}{#2}}

We now evaluate whether \methodname{} turns VLMs into effective high-level reasoners for steering the low-level manipulation policy via instructions. Concretely, we organize the experimental evaluation around the following questions: 
\textbf{(1)} How do different variants of \methodname{} perform on in-distribution and out-of-distribution tasks, and how does \methodname{} compare with instruction-only imitation learning baselines? 
\textbf{(2)} Are the gains from \methodname{} merely due to better representations learned from reasoning supervision, or does explicit test-time compute provide additional benefits?
\textbf{(3)} What specific reasoning behaviors does \methodname{} learn, and how do mid-training and RL change the reasoning traces and induced robot behaviors? We provide a comprehensive set of experiments to answer these questions, alongside comparisons with adaptation of approaches from prior work in our setting.

\begin{table}[t]
    \centering
    \footnotesize
    \setlength{\tabcolsep}{0.7pt}
    \renewcommand{\arraystretch}{1.02}
    
    \begin{tabular}{@{}l *{3}{C{1.28cm}}|*{3}{C{1.28cm}} C{1.5cm} C{1.28cm}|C{1.4cm}|C{1.28cm}@{}}
    \toprule
    &
    \multicolumn{3}{c|}{\textbf{Imitation}} &
    \multicolumn{5}{c|}{\textbf{Ours}} &
    {\Large\bfseries $\Delta$} &
    \textbf{Ref.} \\
    \cmidrule(lr){2-4}\cmidrule(lr){5-9}\cmidrule(lr){10-10}\cmidrule(l){11-11}
    & \makecell[c]{\footnotesize\textbf{Base}\\[-0.15em]\tiny (w/o reason)}
    & \makecell[c]{\footnotesize\textbf{\IL}\\[-0.15em]\tiny (mid only)}
    & {\footnotesize\textbf{\IL}}
    & {\footnotesize\textbf{Base}}
    & \makecell[c]{\footnotesize\textbf{\ours}\\[-0.15em]\tiny (mid only)}
    & \makecell[c]{\footnotesize\textbf{\ours}\\[-0.15em]\tiny (RL only)}
    & \makecell[c]{\footnotesize\textbf{\ours}\\[-0.15em]\tiny (1/4th mid)}
    & {\footnotesize\textbf{\ours}}
    & {\scriptsize \ours{} -- \IL}
    & {\footnotesize\textbf{Gemini}} \\

    \midrule
    \multicolumn{11}{@{}l}{\taskSetMidTrain~~(mid-training tasks)} \\
    \ind{\taskgroup}
    & \valci{11.9}{1.9} & \valci{54.4}{2.9} & \valci{\second{64.7}}{2.8} & \valci{24.1}{2.5} & \valci{53.8}{2.9} & \valci{38.9}{3.3} & \valci{55.3}{2.9} & \valci{\best{65.8}}{3.0} & \valci{$+$1.1}{4.1} & 71.3 \\
    \ind{\taskline}
    & \valci{5.7}{1.4} & \valci{23.6}{2.6} & \valci{\best{33.9}}{2.9} & \valci{8.8}{1.7} & \valci{22.9}{2.5} & \valci{19.4}{2.7} & \valci{31.8}{1.9} & \valci{\second{32.4}}{3.1} & \valci{$-$1.5}{4.2} & 51.0 \\
    \ind{\taskV}
    & \valci{22.8}{2.5} & \valci{25.3}{2.6} & \valci{40.9}{3.0} & \valci{21.1}{2.4} & \valci{33.3}{2.8} & \valci{37.4}{3.2} & \valci{\best{69.4}}{2.6} & \valci{\second{69.2}}{2.9} & \deltaposstrong{28.3}{4.2} & 57.6 \\
    \ind{\taskL}
    & \valci{24.8}{2.6} & \valci{29.6}{2.7} & \valci{\second{34.9}}{2.8} & \valci{25.2}{2.6} & \valci{28.1}{2.7} & \valci{27.6}{3.1} & \valci{29.4}{2.7} & \valci{\best{35.2}}{3.4} & \valci{$+$0.3}{4.4} & 36.2 \\
    \ind{\taskclearqtr}
    & \valci{35.9}{2.7} & \valci{89.0}{1.7} & \valci{91.8}{1.5} & \valci{49.2}{2.7} & \valci{90.7}{1.6} & \valci{86.2}{2.2} & \valci{\second{93.0}}{2.3} & \valci{\best{93.8}}{2.0} & \valci{$+$2.0}{2.5} & 88.9 \\
    \ind{\taskiip}
    & \valci{0.2}{0.3} & \valci{30.8}{2.6} & \valci{\best{35.6}}{2.8} & \valci{1.5}{0.7} & \valci{28.9}{2.6} & \valci{25.3}{2.9} & \valci{28.0}{2.5} & \valci{\second{35.5}}{2.7} & \valci{$-$0.1}{3.9} & 34.0 \\
    \midrule
    
    \multicolumn{11}{@{}l}{\taskSetPostTrain~~(RL tasks)} \\
    \ind{\taskT}
    & \valci{6.4}{1.4} & \valci{6.9}{1.5} & \valci{7.9}{1.6} & \valci{6.2}{1.4} & \valci{6.9}{1.5} & \valci{8.7}{2.0} & \valci{\second{9.0}}{1.4} & \valci{\best{9.9}}{2.2} & \valci{$+$2.0}{2.7} & 10.7 \\
    \ind{\taskgris}
    & \valci{15.0}{2.0} & \valci{27.3}{2.6} & \valci{\best{58.1}}{2.9} & \valci{15.6}{2.1} & \valci{34.8}{2.8} & \valci{24.0}{2.9} & \valci{31.4}{2.3} & \valci{\second{47.8}}{3.3} & \deltanegstrong{10.3}{4.4} & 64.7 \\
    \ind{\taskiV}
    & \valci{19.0}{2.4} & \valci{21.7}{2.5} & \valci{38.9}{2.9} & \valci{17.2}{2.2} & \valci{21.7}{2.5} & \valci{36.9}{3.2} & \valci{\best{61.9}}{2.5} & \valci{\second{57.5}}{2.6} & \deltaposstrong{18.6}{3.9} & 57.0 \\
    \midrule
    
    \multicolumn{11}{@{}l}{\taskSetOOD~~(OOD held-out tasks)}\\
    \ind{\taskdiagline}
    & \valci{23.8}{2.5} & \valci{18.2}{2.3} & \valci{16.7}{2.2} & \valci{\second{34.6}}{2.9} & \valci{\best{37.8}}{2.9} & \valci{26.3}{3.0} & \valci{34.2}{2.8} & \valci{30.9}{3.0} & \deltaposstrong{14.2}{3.7} & 29.9 \\
    \ind{\taskrect}
    & \valci{1.8}{0.8} & \valci{2.1}{0.9} & \valci{2.1}{0.9} & \valci{1.8}{0.8} & \valci{1.7}{0.8} & \valci{\best{12.0}}{2.3} & \valci{\second{6.6}}{1.1} & \valci{6.0}{1.3} & \deltapos{3.9}{1.6} & 9.4 \\
    \ind{\taskmid}
    & \valci{\best{56.3}}{2.8} & \valci{36.4}{2.8} & \valci{42.3}{2.9} & \valci{49.3}{3.0} & \valci{41.7}{2.9} & \valci{\second{53.9}}{3.3} & \valci{45.7}{2.8} & \valci{51.0}{4.1} & \deltapos{8.7}{5.0} & 69.5 \\
    \ind{\taskiL}
    & \valci{23.3}{2.6} & \valci{26.7}{2.7} & \valci{27.3}{2.7} & \valci{25.6}{2.6} & \valci{27.6}{2.7} & \valci{29.2}{3.1} & \valci{\second{30.6}}{2.7} & \valci{\best{37.2}}{3.7} & \deltapos{9.9}{4.6} & 34.4 \\
    \ind{\taskclearhalf}
    & \valci{16.9}{2.0} & \valci{63.1}{2.5} & \valci{69.7}{2.4} & \valci{27.3}{2.3} & \valci{65.6}{2.5} & \valci{59.0}{2.9} & \valci{\best{79.2}}{2.6} & \valci{\second{74.5}}{2.1} & \deltapos{4.8}{3.2} & 74.2 \\
    \bottomrule
    \end{tabular}%
    
    \vspace{-0.2cm}
    \caption{\footnotesize{\textbf{Main results.} We compare base models, imitation baselines, and \methodname{} variants. Values are percentages with 95\% confidence intervals. 
    The \textcolor{green!50!black}{green}/\textcolor{red!70!black}{red} cells in the $\Delta$ column mark significant gains/losses ($|\Delta|>$ CI). Gemini's performance during data collection is shown for reference. \textbf{Bold}/\underline{underlined} values mark the best/second-best non-expert model.}}
    \label{tab:per_task_results}
    \vspace{-0.3cm}
    \end{table}

\parheading{Experimental setup and task design.}
We design the task suite to test whether the learned reasoning aids manipulation. To do so, we construct 14 long-horizon tasks in Language Table, each specified by a high-level textual goal requiring the agent to arrange 8 blocks into spatial patterns. We split these tasks into three splits used for mid-training, RL, and evaluation. The split is chosen to probe three kinds of generalization: \textbf{(i)} transfer to structurally related held-out tasks, \textbf{(ii)} compositional reuse of skills, and \textbf{(iii)} scaling to more difficult geometric arrangements. In particular, we use 6 mid-training tasks 
\(\taskSetMidTrain = \{\taskgroup, \taskline, \taskV, \taskL, \taskclearqtr, \taskiip\}\),
3 additional RL tasks 
\(\taskSetPostTrain = \{\taskT, \taskgris, \taskiV\}\),
and 5 out-of-distribution held-out tasks 
\(\taskSetOOD = \{\taskdiagline, \taskrect, \taskmid, \taskiL, \taskclearhalf\}\).
RL uses all 9 tasks in \(\taskSetMidTrain \cup \taskSetPostTrain\), while \(\taskSetOOD\) is held out for evaluation. Many held-out tasks share structure with training tasks, such as task pairs \((\taskiV,\taskV)\), \((\taskiL,\taskL)\), and \((\taskclearqtr,\taskclearhalf)\); \(\taskgris\) combines skills from \(\taskgroup\) and \(\taskiip\); \(\taskL,\taskT,\taskrect\) form a progression of increasing geometric difficulty. Detailed task descriptions are provided in Appendix~\ref{app:exp_details}.
For each mid-training task, we prompt the expert, Gemini 3 Flash, to collect 4 trajectories per scene over 64 different scenes, resulting in 256 trajectories per task for mid-training. As discussed in Section~\ref{sec:recipe}, we use all collected trajectories for mid-training, rather than only successful ones. For RL, we use 128 successful expert trajectories without reasoning traces per task.

\parheading{Comparisons and evaluation protocol.} We use Qwen3.5-4B as the base model for training. For each task, we evaluate each method on 64 \emph{held-out} scenes with 16 trials per scene and report the average success rate. 
We present main results in Section~\ref{subsec:main_results}, where we compare \methodname{} against two baseline groups: \textbf{(1) recipe ablations}, including mid-training only (\oursMidOnly), RL without mid-training (\oursRLOnly), and RL with mid-training on 1/4th data (\oursQuarterMid); 
\textbf{(2) imitation learning}, i.e., instruction-only SFT without reasoning, using either mid-training data (\ILMid) or all data (\IL).
Additionally, we compare with variants of Embodied Chain-of-Thought (ECoT) reasoning in Section~\ref{subsec:ecot}, where free-form language reasoning is augmented with structured information including end-effector and object states.

\vspace{-0.2cm}
\subsection{Main Performance Results}
\label{subsec:main_results}

\parhighlight{Result 1: RL post-training alone improves performance.}
Table~\ref{tab:per_task_results} reports the per-task success rates across \methodname{} variants and instruction-only imitation baselines. Comparing the base model with \oursRLOnly, we find that RL alone can substantially improve performance on training tasks in \taskSetMidTrain and \taskSetPostTrain. On OOD tasks \taskSetOOD, \oursRLOnly{} also outperforms the base model on all tasks except \taskdiagline. Thus, RL alone can improve task performance by reinforcing useful reasoning behaviors even without mid-training. 

\parhighlight{Result 2: Mid-training improves RL post-training.}
As shown in Table~\ref{tab:per_task_results}, \ours{} outperforms the base model across all mid-training and RL tasks, and consistently improves over \oursRLOnly. On OOD tasks, the gains are more structured: \oursMidOnly~improves on \taskiL~and \taskclearhalf, which are closely related to \taskL~ and \taskclearqtr, but not on \taskrect~or \taskmid; this trend persists after RL. 
The main exception is \taskdiagline, where both \IL~and \ours~degrade performance due to a substantial behavioral gap between the base model and the expert. As shown in Section~\ref{subsec:properties}, they prefer different types of instructions on this task, causing \IL~and \ours~to shift toward behaviors that hurt performance.
Interestingly, although \oursQuarterMid~slightly underperforms full \ours~on \taskSetMidTrain, it already matches or exceeds full \methodname~on \taskSetPostTrain~and \taskSetOOD~. Together, these suggest that mid-training remains a strong warm start, while a modest amount of reasoning data may be sufficient to recover much of the benefit of RL, particularly for OOD generalization.

\parhighlight{Result 3: \methodname~enables better OOD generalization than instruction-only imitation.} As shown in Table~\ref{tab:per_task_results}, \ours~achieves superior or comparable performance to \IL~on both mid-training and post-training tasks. We further emphasize that the main advantage of \ours~lies in its generalization performance. In particular, \ours~outperforms \IL~by a large margin across all five OOD tasks. In contrast, \IL~yields only minor gains or even degrades performance over the base model on OOD tasks except for \taskclearhalf. Notably, on \taskdiagline~and \taskmid, where \IL~hurts base-model performance, \ours~better preserves or improves upon its reasoning base (smaller drop on \taskdiagline~and improvement on \taskmid) while outperforming \IL~on both. These results suggest that \IL~generalizes badly since it primarily memorizes strategies for in-distribution training data, whereas \ours~effectively generalizes the learned reasoning behavior to OOD tasks. 

\begin{AIbox}{Takeaways: \methodname{} generalizes beyond the training distribution}
\begin{itemize}[leftmargin=*, itemsep=1pt, topsep=2pt]
    \item \methodname{} significantly outperforms instruction-only imitation on every held-out OOD task.
    \item RL discovers useful reasoning automatically from instruction-only demonstrations, while mid-training makes that optimization more reliable by providing a strong behavioral prior.
\end{itemize}
\end{AIbox}

\vspace{-0.2cm}
\subsection{Inference-Time Reasoning Matters Beyond Representation Learning}
\label{subsec:representation_learning}
Some prior work has studied the underlying reasons why chain-of-thought reasoning helps robot manipulation, and has primarily concluded that its benefits arise from improved representation learning~\citep{chen2025training,zawalski2025robotic}. One possible reason is that this line of work explicitly constructs reasoning traces that include vision-centric information, such as bounding boxes or object coordinates. In fact, \citet{chen2025training} show that using reasoning at test time is not essential, and that its primary benefit comes from providing additional training-time supervision. These findings somewhat contrast with those in LLMs, where spending additional test-time compute itself leads to improved performance.
A natural question is: are the gains of \ours~explained by better representations learned from reasoning supervision, or does explicit test-time reasoning itself improve generalization? 
We answer this question with three complementary pieces of evidence.
First, we evaluate the models on a VQA suite that probes static perception ability and action-oriented reasoning.
Second, we compare \ours~against instruction-only imitation baselines that receive reasoning supervision via pre-training or co-training, but do not generate test-time reasoning.
Third, we intervene directly on the test-time reasoning budget of the same checkpoint by truncating or removing its reasoning.
Together, these results serve as evidence that \emph{explicit inference-time reasoning improves generalization beyond what is achieved by using reasoning only as training-time supervision.} 

\begin{table}[t]
\centering
\footnotesize
\setlength{\tabcolsep}{3.5pt}
\renewcommand{\arraystretch}{1.02}

\begin{tabular}{@{}l *{7}{C{1.55cm}} @{}}
\toprule
&
\multicolumn{3}{c}{\textbf{Imitation}} &
\multicolumn{4}{c}{\textbf{Ours}} \\
\cmidrule(lr){2-4}\cmidrule(l){5-8}
& {\footnotesize\textbf{\IL}}
& \makecell[c]{\footnotesize\textbf{\IL}\\[-0.15em]\tiny (Pre-train)}
& \makecell[c]{\footnotesize\textbf{\IL}\\[-0.15em]\tiny (Co-train)}
& {\footnotesize\textbf{\ours}}
& \makecell[c]{\footnotesize\textbf{\ours}\\[-0.15em]\tiny (trunc@100)}
& \makecell[c]{\footnotesize\textbf{\ours}\\[-0.15em]\tiny (trunc@50)}
& \makecell[c]{\footnotesize\textbf{\ours}\\[-0.15em]\tiny (w/o reason)} \\

\midrule
\multicolumn{8}{@{}l}{\taskSetMidTrain~~(mid-training tasks)} \\
\ind{\taskgroup} & 64.7 & 65.9\,\tabdlt{+1.2} & 66.5\,\tabdlt{+1.8} & 65.8 & 60.9\,\tabdlt{-4.9} & 53.1\,\tabdlt{-12.7} & 39.8\,\tabdlt{-26.0} \\
\ind{\taskline} & 33.9 & 34.6\,\tabdlt{+0.7} & 33.9\,\tabdlt{0.0} & 32.4 & 28.9\,\tabdlt{-3.5} & 21.1\,\tabdlt{-11.3} & 17.6\,\tabdlt{-14.8} \\
\ind{\taskV} & 40.9 & 43.4\,\tabdlt{+2.5} & 36.6\,\tabdlt{-4.3} & 69.2 & 59.4\,\tabdlt{-9.8} & 26.6\,\tabdlt{-42.6} & 20.3\,\tabdlt{-48.9} \\
\ind{\taskL} & 34.9 & 38.3\,\tabdlt{+3.4} & 35.4\,\tabdlt{+0.5} & 35.2 & 30.5\,\tabdlt{-4.7} & 30.5\,\tabdlt{-4.7} & 32.0\,\tabdlt{-3.2} \\
\ind{\taskclearqtr} & 91.8 & 91.8\,\tabdlt{0.0} & 92.1\,\tabdlt{+0.3} & 93.8 & 96.9\,\tabdlt{+3.1} & 91.4\,\tabdlt{-2.4} & 91.0\,\tabdlt{-2.8} \\
\ind{\taskiip} & 35.6 & 33.6\,\tabdlt{-2.0} & 33.6\,\tabdlt{-2.0} & 35.5 & 35.2\,\tabdlt{-0.3} & 35.9\,\tabdlt{+0.4} & 24.2\,\tabdlt{-11.3} \\
\midrule

\multicolumn{8}{@{}l}{\taskSetPostTrain~~(RL tasks)} \\
\ind{\taskT} & 7.9 & 7.8\,\tabdlt{-0.1} & 8.6\,\tabdlt{+0.7} & 9.9 & 9.0\,\tabdlt{-0.9} & 7.0\,\tabdlt{-2.9} & 7.4\,\tabdlt{-2.5} \\
\ind{\taskgris} & 58.1 & 60.8\,\tabdlt{+2.7} & 50.3\,\tabdlt{-7.8} & 47.8 & 44.1\,\tabdlt{-3.7} & 25.0\,\tabdlt{-22.8} & 22.1\,\tabdlt{-25.7} \\
\ind{\taskiV} & 38.9 & 48.2\,\tabdlt{+9.3} & 37.9\,\tabdlt{-1.0} & 57.5 & 52.3\,\tabdlt{-5.2} & 27.3\,\tabdlt{-30.2} & 17.6\,\tabdlt{-39.9} \\
\midrule

\multicolumn{8}{@{}l}{\taskSetOOD~~(OOD held-out tasks)} \\
\ind{\taskdiagline} & 16.7 & 21.9\,\tabdlt{+5.2} & 25.1\,\tabdlt{+8.4} & 30.9 & 28.1\,\tabdlt{-2.8} & 24.2\,\tabdlt{-6.7} & 21.9\,\tabdlt{-9.0} \\
\ind{\taskrect} & 2.1 & 1.8\,\tabdlt{-0.3} & 1.4\,\tabdlt{-0.7} & 6.0 & 3.9\,\tabdlt{-2.1} & 4.7\,\tabdlt{-1.3} & 2.3\,\tabdlt{-3.7} \\
\ind{\taskmid} & 42.3 & 44.1\,\tabdlt{+1.8} & 38.9\,\tabdlt{-3.4} & 51.0 & 52.0\,\tabdlt{+1.0} & 43.0\,\tabdlt{-8.0} & 50.0\,\tabdlt{-1.0} \\
\ind{\taskiL} & 27.3 & 33.3\,\tabdlt{+6.0} & 29.3\,\tabdlt{+2.0} & 37.2 & 37.9\,\tabdlt{+0.7} & 30.5\,\tabdlt{-6.7} & 27.7\,\tabdlt{-9.5} \\
\ind{\taskclearhalf} & 69.7 & 69.0\,\tabdlt{-0.7} & 74.4\,\tabdlt{+4.7} & 74.5 & 73.4\,\tabdlt{-1.1} & 73.8\,\tabdlt{-0.7} & 76.0\,\tabdlt{+1.5} \\
\bottomrule
\end{tabular}

\vspace{-0.2cm}
\caption{\footnotesize{\textbf{\IL~with pre-training or co-training on reasoning, and truncation of \ours{} reasoning at test time.} Values are percentages. Parenthetical values for \ILPretrain{} and \ILCotrain{} indicate absolute changes relative to \IL{}; for truncated \ours{} variants, they indicate absolute changes relative to full \ours{}. 
}}
\label{tab:pretrain_cotrain_results}
\vspace{-0.45cm}
\end{table}

\parhighlight{Evidence A: \methodname{} improves both static perception and action understanding, but these improvements alone do not explain its manipulation gains.}
To understand what the trained VLM reasoner learns, we evaluate models on a visual question-answering (VQA) suite. The suite probes both static perception, such as object localization and spatial relations, and action-oriented reasoning, such as inferring the instruction that would produce a given transition. We provide details of VQA tasks in Appendix~\ref{app:subsec:vqa_results} and results in Table~\ref{app:tab:vqa_results}. Overall, \emph{we find that both mid-training and RL improve VQA performance.} The gains are small on simple absolute-position questions, but larger on relative-position and distance questions, which require understanding relationships of multiple blocks. \methodname{} also substantially improves instruction inference, suggesting that reasoning training also improves the model's ability to connect scene states to appropriate high-level actions. In contrast, instruction-execution questions improve less, likely because they require judging fine-grained success criteria for instructions, which is not optimized in training.

These diagnostics suggest that \methodname{} improves both static perception and action understanding, but they also show that VQA performance alone, i.e., improving static perception, does not fully explain the gains in manipulation performance. Even our best model remains far below Gemini on several VQA categories, yet matches or approaches Gemini on many manipulation tasks. Thus, \emph{the gains from \methodname{} are not simply due to better static perception; they likely also come from changes in how the model uses language reasoning to steer the low-level policy.} 

\parhighlight{Evidence B: \ours{} generalizes better than non-reasoning policies that use reasoning as additional training-time supervision.}
Following~\citet{chen2025training}, we test whether the benefits of reasoning can be absorbed into an instruction-only imitation-learning policy through training-time supervision alone. To do so, we incorporate reasoning-labeled examples into the imitation-learning baseline in two ways, while removing reasoning at test time. In the pre-training variant, we initialize imitation learning from our mid-trained reasoner, \oursMidOnly, rather than from the base Qwen3.5-4B model. In the co-training variant, we train on a mixture of the reasoning-labeled mid-training data used by \methodname{} and the instruction-only imitation-learning data. 
We denote these variants as \ILPretrain{} and \ILCotrain{}, respectively.

Table~\ref{tab:pretrain_cotrain_results} compares the results of \IL, \ILPretrain, \ILCotrain, and \ours. On mid-training tasks, the four models achieve broadly comparable performance, with the notable exception of \taskV, where \ours~substantially outperforms the imitation learning variants. On tasks unseen during mid-training (\taskSetPostTrain~and \taskSetOOD), adding reasoning supervision through pre-training or co-training improves generalization in several cases: pre-training yields sizable gains on \taskiV~and \taskiL, while co-training improves performance on \taskdiagline~and \taskclearhalf. These indicate that reasoning supervision can indeed help imitation learning by improving representations. However, these gains are not sufficient to match the performance of our approach. In particular, on OOD tasks, \ours~consistently outperforms all imitation variants. \begin{wrapfigure}{r}{0.3\textwidth}
    \centering
    \vspace{-0.2cm}
    \includegraphics[width=\linewidth]{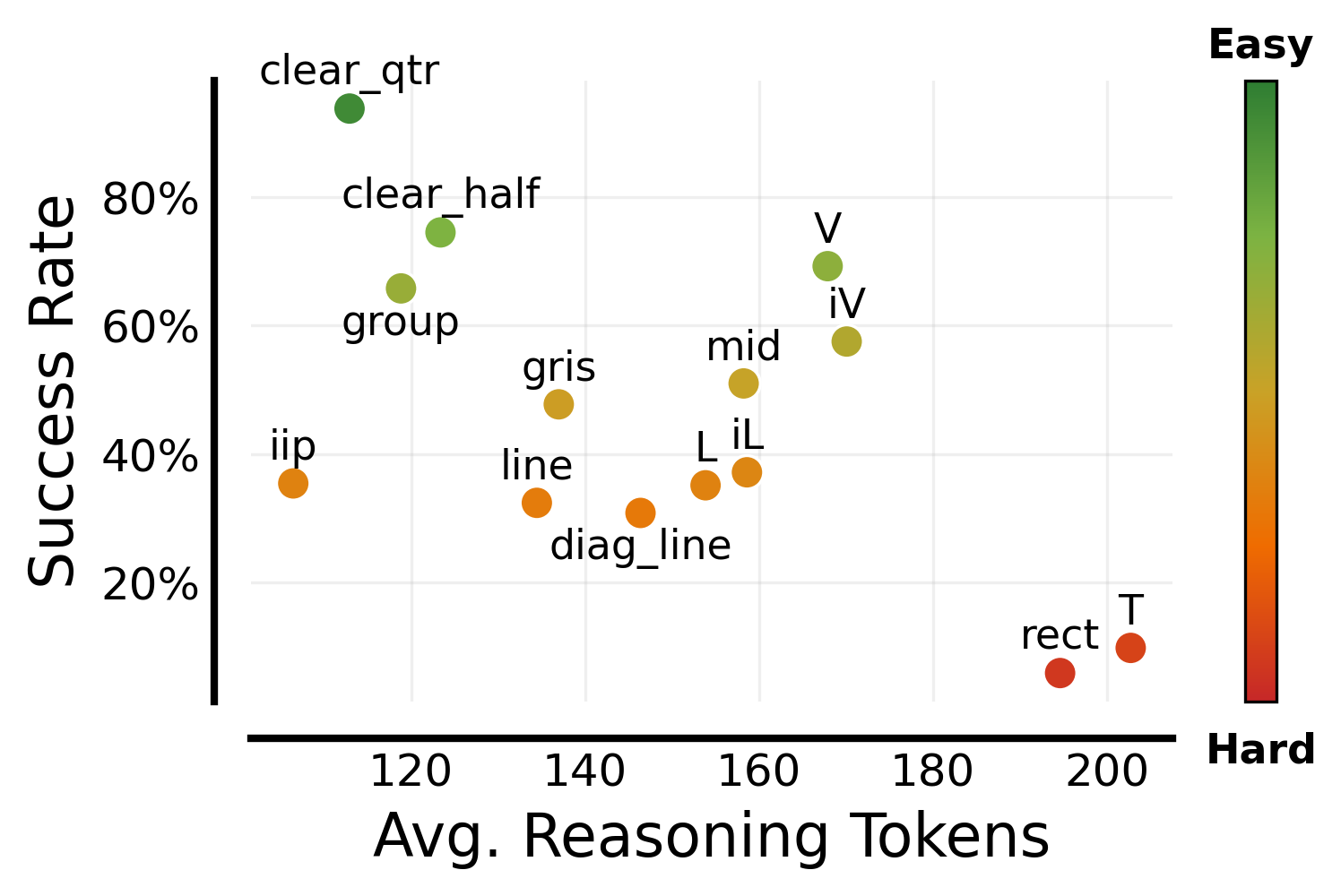}
    \vspace{-0.65cm}
    \caption{\footnotesize{\textbf{Per-task token length.}}}
    \label{fig:success_vs_tokens}
    \vspace{-0.6cm}
\end{wrapfigure}
\emph{This gap suggests that reasoning cannot simply be discarded at inference time; the benefit of reasoning cannot be fully captured by pre-training} Instead, explicit inference-time reasoning provides an additional source of generalization by enabling the model to plan, adapt, and select task-relevant instructions online.

\parhighlight{Evidence C: Increasing the inference-time reasoning budget improves success.}
Using the same \ours~checkpoint, we vary the reasoning budget to compare performance using no reasoning, reasoning truncated at 50 or 100 tokens, and full reasoning.
In Table~\ref{tab:pretrain_cotrain_results} (``Ours'' group), we see that allowing a larger reasoning budget generally yields marked gains, especially on \taskgroup{}, \taskline{}, \taskV{}, \taskiip{}, \taskgris{}, \taskiV{}, \taskdiagline{}, \taskrect{}, and \taskiL{}.
Since these variants differ only in the inference-time reasoning budget, this comparison isolates the effect of inference-time reasoning while holding the learned representations fixed.
Additionally, Figure~\ref{fig:success_vs_tokens} shows that our model generally elicits longer reasoning traces on harder, lower-success tasks.
Together, these results provide causal evidence that reasoning contributes substantially to task success and serves as useful test-time compute.

\begin{AIbox}{Takeaways: Reasoning and representation learning}
\begin{itemize}[leftmargin=*, itemsep=1pt, topsep=2pt]
    \item While \methodname{} does improve perception and action understanding, gains in perception alone do not explain improvements on reasoning and manipulation.
    \item Using reasoning data as auxiliary supervision for co-training does not explain its benefits.
    \item Increasing the inference-time budget improves performance and the trained reasoner spends more tokens for reasoning on harder tasks.
\end{itemize}
\end{AIbox}

\vspace{-0.2cm}
\subsection{Understanding Reasoning Behaviors Learned by \methodname}
\label{subsec:properties}

\begin{wrapfigure}{r}{0.45\textwidth}
    \centering
    \vspace{-0.2cm}
    \includegraphics[width=\linewidth]{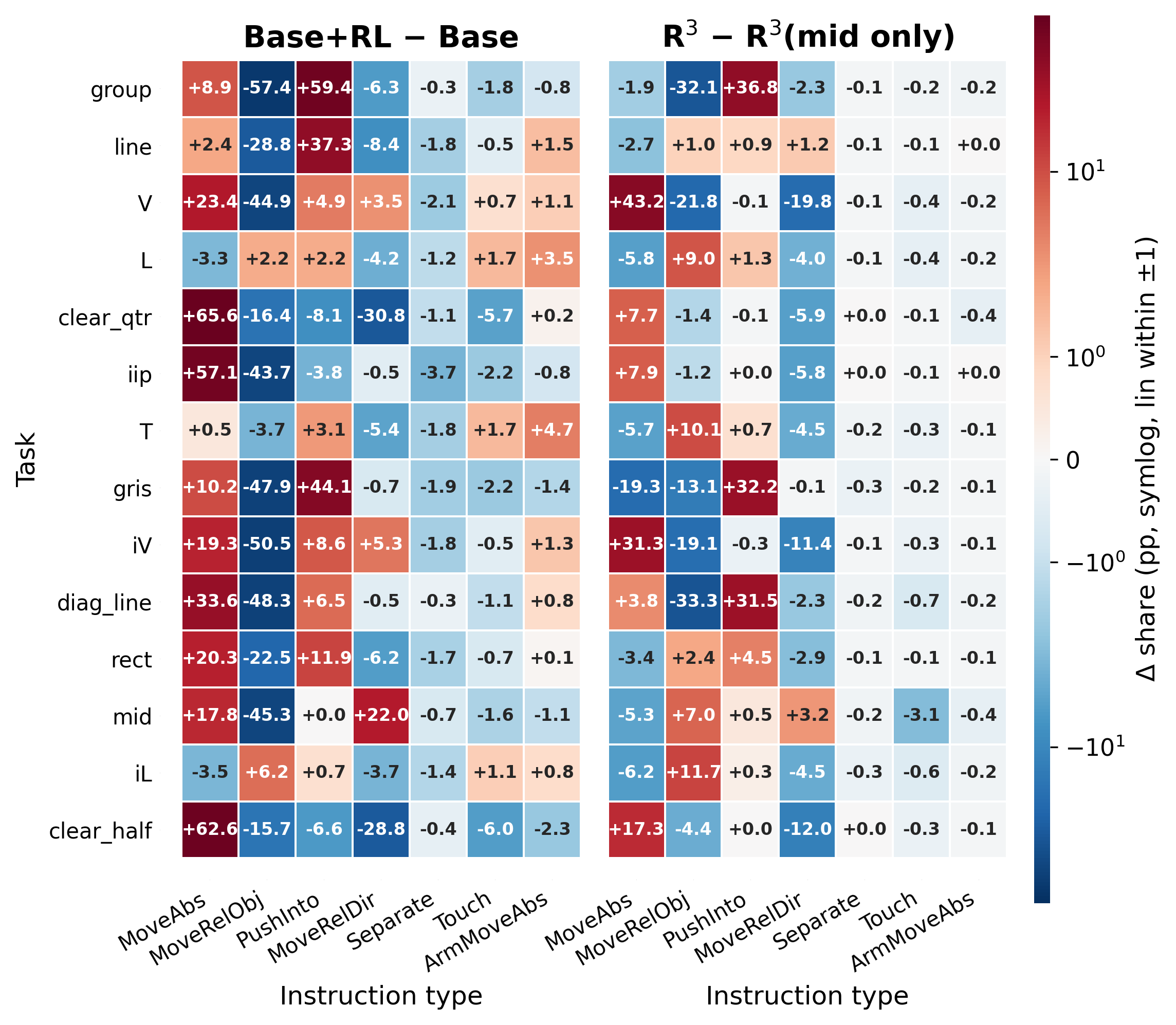}
    \vspace{-0.5cm}
    \caption{\footnotesize{\textbf{RL affects instruction distributions.} RL from the base model broadly rewrites the instruction distribution, while RL after mid-training makes localized edits.}}
    \vspace{-0.2cm}
    \label{fig:model_instr_distribution}
\end{wrapfigure}
\parhighlight{\methodname{} learns reasoning strategies useful for manipulation.}
Beyond success rates, we qualitatively inspect the reasoning traces produced by \methodname~ in Appendix~\ref{app:examples}. These traces suggest that \methodname~learns behaviors useful for long-horizon manipulation that are under-represented in the mid-training data. First, the trained reasoner \emph{compares multiple alternatives and performs self-correction} before choosing instructions. In Figure~\ref{fig:reasoning_example_1_backtrack}, it considers object-vertex assignments, notices that its initial plan is inconsistent with the current scene, and revises the plan. Second, the reasoner uses reasoning to \emph{resolve visual and historical uncertainty}. In Figure~\ref{fig:reasoning_example_2_occlusion}, an object is partially occluded by the arm; rather than blindly following the previous response, the model re-examines the scene, task information, and history to infer the correct object state.
We further qualitatively compare aligned reasoning traces from the base, mid-trained, and RL-trained variants on the same 30 validation scenes. Full analyses are in Appendix~\ref{app:reasoning_trace_analysis}. The base model is exploratory but unreliable, with variable formatting, hallucinated objects, non-convergent backtracking, and malformed instructions. Mid-training largely fixes these interface-level failures, but often gives terse and single-pass reasoning. RL after mid-training preserves this reliability while making reasoning more state-aware: the model more often restates task constraints, tracks progress, and chooses next steps incrementally. Together, these comparisons suggest that mid-training stabilizes the reasoning interface, and RL refines it into more deliberate action-oriented planning.

\parhighlight{Mid-training learns behavioral priors aligned with the expert that RL selectively updates.}
We analyze the distribution of instruction primitives produced by each model in Figure~\ref{fig:model_instr_distribution}. 
The base model's distribution differs substantially from the expert's, overusing short-horizon primitives such as \texttt{MoveRelObj} and underusing \texttt{MoveAbs} on spatial-target tasks. 
RL from the base model improves success, but does not recover the expert distribution; instead, it often shifts the policy toward a different mode, overusing \texttt{MoveAbs} or \texttt{PushInto}.
Mid-training largely aligns the model's instruction distribution with the expert's, providing a strong behavioral prior before RL. When RL is applied after mid-training, the resulting instruction distribution remains largely consistent with the mid-trained model, with larger shifts occurring mainly where the mid-trained model is still mismatched with the expert, such as increased use of \texttt{PushInto} on \taskdiagline{}/\taskgris{} and \texttt{MoveAbs} on \taskV{} and \taskiV{}. This supports the role of mid-training as a warm start: from a weak prior, RL must discover new useful behaviors, whereas \emph{with mid-training, RL can refine an already reasonable behavior distribution}.
This refinement also reflects the \emph{mode-seeking} nature of RL, illustrated by the example in Figure~\ref{fig:V_traj_comparison}.
For such tasks where expert strategies are diverse, mid-training exposes the model to a broad range of behaviors, and subsequent RL tends to concentrate its behavior around good strategies for higher rewards. 
We also observe that rare instructions (\texttt{Separate}, \texttt{Touch}, and \texttt{ArmMoveAbs}) collapse after RL, suggesting that behaviors misaligned with the expert are dropped in RL.

\vspace{-0.2cm}
\subsection{Comparison with Approaches that Use Structured CoT Templates}
\label{subsec:ecot}

Prior works often rely on structured reasoning templates,
which encode procedural reasoning patterns shared across scenes. For example, ECoT~\citep{zawalski2025robotic,chen2025training} constructs a vision-centric CoT containing bounding boxes and object coordinates. Related approaches structure intermediate reasoning through object, grasp, and affordances~\citep{li2025coavla}, visual subgoals~\citep{zhao2025cotvla}, end-effector paths~\citep{li2025hamster}, depth-aware perception and image-space trajectories~\citep{lee2025molmoact}, or structured descriptions of driving states~\citep{gao2026steervlasteeringvisionlanguageactionmodels}. In this section, we compare our approach against an adaptation of ECoT to our setting, where we augment the reasoning trace with explicit annotations obtained from simulator state. Specifically, our ECoT implementation includes the task goal, end-effector state, object states, language reasoning, and the resulting instruction. 

\begin{wraptable}{r}{0.35\textwidth}
\centering
\vspace{-0.3cm}
\scriptsize
\setlength{\tabcolsep}{2.3pt}
\renewcommand{\arraystretch}{1.02}
\resizebox{\linewidth}{!}{%
\begin{tabular}{@{}l *{4}{c} @{}}
\toprule
& {\footnotesize\textbf{Ours}}
& {\footnotesize\textbf{Post-hoc}}
& {\footnotesize\textbf{Ours}}
& {\footnotesize\textbf{Post-hoc}} \\
& & & {\footnotesize\textbf{+ ECoT}} & {\footnotesize\textbf{+ ECoT}} \\
\midrule
\multicolumn{5}{@{}l}{\taskSetMidTrain~~(mid-training tasks)} \\
\ind{\taskgroup} & 53.8 & \best{53.9} & 51.0 & 46.6 \\
\ind{\taskline} & \best{22.9} & 19.7 & 20.6 & 20.6 \\
\ind{\taskV} & \best{33.3} & 28.7 & 28.4 & 30.2 \\
\ind{\taskL} & 28.1 & \best{31.1} & 28.3 & 25.9 \\
\ind{\taskclearqtr} & \best{90.7} & 89.3 & 87.5 & 86.7 \\
\ind{\taskiip} & \best{28.9} & 27.7 & 27.9 & 24.7 \\
\midrule
\multicolumn{5}{@{}l}{\taskSetPostTrain~and \taskSetOOD~~(held-out tasks)} \\
\ind{\taskT} & 6.9 & \best{10.3} & 6.6 & 7.4 \\
\ind{\taskgris} & \best{34.8} & 24.6 & 26.4 & 25.8 \\
\ind{\taskiV} & 21.7 & \best{22.8} & 21.9 & 22.4 \\
\ind{\taskdiagline} & \best{37.8} & 35.2 & 36.6 & 34.8 \\
\ind{\taskrect} & 1.7 & 2.3 & 1.7 & \best{2.6} \\
\ind{\taskmid} & \best{41.7} & 39.8 & 39.8 & 36.7 \\
\ind{\taskiL} & 27.6 & 25.9 & 27.2 & \best{28.8} \\
\ind{\taskclearhalf} & 65.6 & \best{66.7} & 66.4 & 62.8 \\
\bottomrule
\end{tabular}
}
\vspace{-0.15cm}
\caption{\footnotesize{
\textbf{Evaluation results of ECoT variants.} Trained by SFT on only our mid-training data. Values are percentages. \textbf{Bold} values mark the best model.
}}
\label{tab:ecot_results}
\vspace{-0.4cm}
\end{wraptable}
We compare four variants of reasoning supervision for mid-training, defined along two axes:
\textbf{(1) when the reasoning annotations are obtained:} reasoning recorded during data collection (``Ours'') versus reasoning retrospectively labeled on demonstration data (``Post-hoc''), following prior ECoT  training~\citep{zawalski2025robotic,chen2025training};
and \textbf{(2) reasoning style:} free-form language reasoning versus ECoT-style reasoning
(``+ECoT'').
We then derive \emph{four} different combinations of SFT data: ``Ours", ``Post-hoc", ``Ours+ECoT", and ``Post-hoc+ECoT".
Note that ``Ours" here refers exactly to ``\oursMidOnly"~for the main results. 
We train all variants by SFT on our mid-training data, which contains only \taskSetMidTrain~tasks, and \taskSetPostTrain~and \taskSetOOD~are used as held-out tasks for evaluation. 
Our motivation and implementation details of ECoT are provided in
Appendix~\ref{app:subsec:ecot}.

\parhighlight{ECoT-style reasoning does not provide additional benefit.}
Table~\ref{tab:ecot_results} reports the evaluation results of the four variants. 
Comparing the ECoT variants with their non-ECoT counterparts, we find that adding ECoT components slightly degrades overall performance. This suggests that \emph{in our setting, ECoT does not provide additional benefit over our free-form language reasoning.} One possible explanation is that our tasks require reasoning about long-horizon task progress, execution failures, and closed-loop replanning, whereas the additional ECoT components primarily make low-level visual grounding information explicit, such as end-effector and object states. 
Comparing ``Ours'' with ``Post-hoc'', we find that post-hoc reasoning performs comparably to reasoning recorded during data collection. In theory, one could expect that reasoning recorded during the data collection process should outperform post-hoc labeling since the former captures the true causal factors behind the action, whereas the latter only provides a potentially imperfect estimate, but they perform comparably in our experiments. 
We hypothesize that this is because our post-hoc reasoning traces are generated by Gemini for trajectories also collected by Gemini. As a result, post-hoc reasoning generation may be easier than in a more realistic setting where the reasoning model must explain actions produced by a different expert, such as human or another policy. A more complete analysis would require systematically varying both the trajectory collector and reasoning generator, e.g., using GPT to generate post-hoc reasoning for Gemini trajectories. We leave this comparison to future work.

\begin{AIbox}{Takeaways: What makes \methodname{} effective}
\begin{itemize}[leftmargin=*, itemsep=1pt, topsep=2pt]
    \item Mid-training anchors a broad, expert-like behavior distribution; RL then refines that prior into more deliberate action-oriented planning and concentrated behaviors.
    \item Reasoning enables progress tracking, recovery from failures, and closed-loop replanning that matters more than visual grounding, making structured CoTs less useful.
\end{itemize}
\end{AIbox}

\vspace{-0.2cm}
\section{Experiments on Bimanual Grocery Packing}
\label{sec:packing}
\vspace{-0.2cm}

\parheading{Experimental setup and task design.}
We next test whether our approach extends to a bimanual grocery packing task from forthcoming work~\citep{anonymous2026vlaexplore}. The task is instantiated in a dual-arm grocery packing workspace using a dual xArm-7 platform, following RaC~\citep{hu2025rac}. Detailed descriptions of the environment setup, task goals, success criteria, and a comparison with Language Table are provided in Appendix~\ref{app:packing} and Appendix~\ref{app:lt_vs_packing}.
We use the dataset of human teleoperation data labeled with instructions directly from \citet{anonymous2026vlaexplore}, and fine-tune \(\pi_{0.5}\)~\citep{intelligence2025pi05visionlanguageactionmodelopenworld} on this data to obtain a steerable low-level policy. 
We skip Stage I mid-training because the base VLM already produces useful reasoning on this domain and because no reasoning annotations are provided in \citet{anonymous2026vlaexplore}. 
To construct data for Stage II RL, we sample frames from each segment, oversampling the onset and pre-completion of the subtask. See Appendix~\ref{app:packing} for details. Each training example conditions the VLM on the three current views and the long-horizon goal and supervises the current instruction. 

\parheading{Comparisons and evaluation protocol.}
We use the same Qwen3.5-4B base model as on Language Table. We compare: \textbf{(1)} the base model with and without reasoning; \textbf{(2)} instruction-only imitation (\IL) without reasoning; and \textbf{(3)} our method \oursRLOnly, i.e., RL with the string-match reward, initialized from the base model.
We evaluate on 12 \emph{held-out} task configurations unseen in the training data, \taskspack{1}--\taskspack{12}. 
For each of the 12 held-out tasks we run 5 environment seeds \(\times\) 10 rollouts (50 episodes per task; 600 in total). An episode is successful if every goal object is stably packed in its assigned tray, designated clutter is cleared, and any orientation constraint is satisfied. We also report normalized task progress as a metric, which reflects the fraction of packing stages (goal objects) completed.

\label{subsec:packing_results}

\begin{table}[t]
    \centering
    \footnotesize
    \setlength{\tabcolsep}{1.2pt}
    \renewcommand{\arraystretch}{1.02}
    \begin{tabular}{@{}l *{4}{C{1.65cm}}|*{4}{C{1.65cm}}@{}}
    \toprule
    &
    \multicolumn{4}{c|}{\textbf{Success Rate}} &
    \multicolumn{4}{c}{\textbf{Progress}} \\
    \cmidrule(lr){2-5}\cmidrule(l){6-9}
    & \makecell[c]{\footnotesize\textbf{Base}\\[-0.15em]\tiny (w/o reason)}
    & \makecell[c]{\footnotesize\textbf{\IL}\\[-0.15em]\tiny (w/o reason)}
    & \makecell[c]{\footnotesize\textbf{Base}\\[-0.15em]\tiny (w/ reason)}
    & {\footnotesize\textbf{Ours}}
    & \makecell[c]{\footnotesize\textbf{Base}\\[-0.15em]\tiny (w/o reason)}
    & \makecell[c]{\footnotesize\textbf{\IL}\\[-0.15em]\tiny (w/o reason)}
    & \makecell[c]{\footnotesize\textbf{Base}\\[-0.15em]\tiny (w/ reason)}
    & {\footnotesize\textbf{Ours}} \\
    \midrule
    \ind{\taskspack{1}}
    & \valci{76.0}{12.0} & \valci{42.0}{13.3} & \valci{84.0}{15.7} & \valci{\best{90.0}}{7.9} & \valci{81.0}{10.1} & \valci{69.0}{7.6} & \valci{86.0}{14.1} & \valci{\best{91.0}}{7.4} \\
    \ind{\taskspack{2}}
    & \valci{\best{100.0}}{0.0} & \valci{\best{100.0}}{0.0} & \valci{96.0}{7.8} & \valci{\best{100.0}}{0.0} & \valci{\best{100.0}}{0.0} & \valci{\best{100.0}}{0.0} & \valci{98.7}{2.6} & \valci{\best{100.0}}{0.0} \\
    \ind{\taskspack{3}}
    & \valci{0.0}{0.0} & \valci{40.0}{14.1} & \valci{0.0}{0.0} & \valci{\best{60.0}}{14.2} & \valci{59.0}{2.7} & \valci{78.0}{5.7} & \valci{38.7}{4.9} & \valci{\best{84.0}}{6.5} \\
    \ind{\taskspack{4}}
    & \valci{4.0}{5.5} & \valci{12.0}{9.2} & \valci{12.0}{13.6} & \valci{\best{26.0}}{12.7} & \valci{48.0}{6.5} & \valci{49.0}{7.2} & \valci{45.0}{10.2} & \valci{\best{66.5}}{7.8} \\
    \ind{\taskspack{5}}
    & \valci{0.0}{0.0} & \valci{0.0}{0.0} & \valci{0.0}{0.0} & \valci{\best{16.0}}{10.3} & \valci{34.0}{4.8} & \valci{31.6}{3.4} & \valci{38.4}{6.2} & \valci{\best{66.0}}{6.5} \\
    \ind{\taskspack{6}}
    & \valci{6.0}{6.8} & \valci{16.0}{10.3} & \valci{0.0}{0.0} & \valci{\best{28.0}}{12.4} & \valci{42.0}{8.2} & \valci{52.5}{9.1} & \valci{43.0}{11.4} & \valci{\best{55.5}}{10.7} \\
    \ind{\taskspack{7}}
    & \valci{0.0}{0.0} & \valci{\best{44.0}}{13.7} & \valci{0.0}{0.0} & \valci{10.7}{15.2} & \valci{50.0}{0.0} & \valci{\best{74.0}}{7.0} & \valci{50.0}{0.0} & \valci{55.3}{7.6} \\
    \ind{\taskspack{8}}
    & \valci{0.0}{0.0} & \valci{0.0}{0.0} & \valci{4.0}{7.8} & \valci{\best{6.0}}{6.8} & \valci{28.8}{5.9} & \valci{21.2}{7.0} & \valci{35.2}{9.7} & \valci{\best{36.0}}{10.5} \\
    \ind{\taskspack{9}}
    & \valci{20.0}{10.9} & \valci{\best{86.0}}{8.6} & \valci{36.0}{18.4} & \valci{84.0}{10.6} & \valci{67.5}{6.7} & \valci{\best{92.0}}{5.7} & \valci{74.0}{10.4} & \valci{91.0}{7.1} \\
    \ind{\taskspack{10}}
    & \valci{10.0}{8.4} & \valci{48.0}{14.0} & \valci{16.0}{14.7} & \valci{\best{54.0}}{14.3} & \valci{51.5}{6.9} & \valci{83.5}{5.4} & \valci{54.0}{12.2} & \valci{\best{84.0}}{6.0} \\
    \ind{\taskspack{11}}
    & \valci{20.0}{11.4} & \valci{54.0}{13.5} & \valci{32.0}{18.4} & \valci{\best{72.0}}{12.8} & \valci{69.6}{6.8} & \valci{84.8}{6.4} & \valci{80.0}{8.8} & \valci{\best{87.2}}{7.1} \\
    \ind{\taskspack{12}}
    & \valci{0.0}{0.0} & \valci{14.0}{9.1} & \valci{0.0}{0.0} & \valci{\best{28.0}}{11.5} & \valci{28.4}{7.2} & \valci{48.8}{9.4} & \valci{36.0}{13.1} & \valci{\best{60.8}}{10.3} \\
    \midrule
    \ind{Mean}
    & \valci{19.7}{1.9} & \valci{38.0}{3.0} & \valci{23.3}{3.2} & \valci{\best{47.9}}{3.3} & \valci{55.0}{1.8} & \valci{65.4}{1.9} & \valci{56.6}{2.8} & \valci{\best{73.1}}{2.2} \\
    \bottomrule
    \end{tabular}
    \caption{\footnotesize{\textbf{Main results on grocery packing.} Success rate and normalized progress on 12 held-out tasks. Values are percentages with 95\% confidence intervals. \textbf{Bold} values mark the best model.}}
    \label{tab:packing_results}
\end{table}

\parhighlight{Result: \oursRLOnly{} outperforms instruction-only imitation.}
Table~\ref{tab:packing_results} reports per-task success rates and normalized progress. Consistent with Language Table, our approach attains substantially higher overall success and progress than instruction-only \IL{} without reasoning. 
These results suggest that our recipe successfully transfers to long-horizon bimanual manipulation beyond Language Table, and that Stage I mid-training can be skipped when the base VLM already produces useful reasoning on the target domain.
Qualitative examples of rollouts in Appendix~\ref{app:examples} show the same action-oriented behaviors as on Language Table: the reasoner tracks packing progress from the three camera views (Figures~\ref{fig:reasoning_example_pack_1_typical} and~\ref{fig:reasoning_example_pack_2_typical}) and can re-examine the scene to correct an initially wrong object localization (Figure~\ref{fig:reasoning_example_pack_3_correction}).

%% file: sections/conclusion.tex
\vspace{-0.2cm}
\section{Discussion and Perspectives on Future Work}
\label{sec:conclusion}
\vspace{-0.2cm}

We introduced \methodname{}, a recipe for training VLMs to reason flexibly in natural language before issuing high-level instructions that steer a fixed low-level robot policy. By combining mid-training on expert reasoning traces with rubric-based single-step reinforcement learning from offline action data, \methodname{} learns to generate action-oriented reasoning that can guide a frozen low-level robot policy. Experiments on Language Table and bimanual grocery packing show that this approach improves performance over instruction-only imitation without reasoning. On Language Table, \methodname{} further improves across both seen and unseen long-horizon tasks and demonstrates stronger out-of-distribution generalization. We further find that the trained reasoner learns behaviors such as tracking interaction history, resolving visual ambiguity, and self-correction. We show that free-form
language reasoning can function as an effective test-time compute mechanism for steering low-level policies.

\parheading{Limitations.} First, our experiments are conducted in two simulated domains (Language Table and bimanual grocery packing) with a fixed low-level language-conditioned policy. While this helps us perform a systematic study, extending the approach to real robots remains important future work. Second, Stage I still relies on expert-generated reasoning, and Stage II relies on a VLM judge to provide semantic rewards. While this reduces the need for multi-turn robot rollouts, it also optimizes a surrogate objective. Extending our approach to multi-turn online RL will further improve long-horizon behaviors.

\parheading{Future work.} 
We believe there are several directions for future work. First, \ours~could be deployed on real robots and evaluated on more dexterous, longer-horizon manipulation tasks. This would test whether natural-language reasoning remains useful under real-world challenges such as noisy perception, physical recovery, and adaptation to unseen environments.
Second, future work could reduce the separation between the high-level reasoner and the low-level robot policy. While our hierarchical design isolates the effect of reasoning, it may introduce a mismatch between high-level intent and low-level execution. Jointly training reasoning and action prediction could improve coordination while preserving the interpretability benefits of language-based reasoning. It would also be interesting to study whether reasoning can support not only high-level steering, but also the generation of low-level actions themselves. The closest existing approaches condition action generation on intermediate metadata~\citep{intelligence2026pi07steerablegeneralistrobotic}, but again rely on highly structured templates, suggesting that substantial gains remain to be realized. In addition, joint training introduces technical systems challenges around synchronizing high-level reasoning with low-level action, which will be important to address.
Third, the RL stage could be extended beyond single-step offline training. Our current formulation avoids expensive online interaction by rewarding semantic agreement with expert instructions, but it optimizes a surrogate objective rather than final task success. Multi-turn RL with feedback from task completion, intermediate progress, recovery behavior, or human preferences may further improve long-horizon reasoning.
Finally, \ours~could be extended to support online improvement. Rather than relying solely on batched offline training, the reasoning VLM could be updated from environment feedback or human corrections. Free-form reasoning may also provide a natural mechanism for exploration, allowing the robot to expand the support of its behavior beyond what is represented in the offline data. While most robot RL methods today focus on sharpening an existing low-level policy, reasoning could enable qualitatively broader exploration, supporting robust adaptation to new embodiments, novel objects, and failure modes not encountered during training.

%% file: sections/acknowledgment.tex
\section*{Acknowledgments}

We thank Kshitiz and Robyn Wu for support with the bimanual grocery packing environment and data from their forthcoming work~\citep{anonymous2026vlaexplore}. We thank Max Sobol Mark, Ian Wu, Kushal Arora, Abhishek Gupta, Marius Memmel and Mateo Castro for informative discussions. We thank members of CMU AIRe and RCHI labs for their support. This work is supported by the Office of Naval Research under N00014-24-12206, a Schmidt Sciences AI2050 Early Career Fellowship, and a TRI U3.0 project. We thank the Orchard cluster at the CMU FLAME center for support with GPU resources and TPU research cloud (TRC) for their support with TPU resources. YQ gratefully acknowledges support from the Amazon AI PhD Fellowship.

%% file: appendix/exp_details.tex
\section{Experimental Details}
\label{app:exp_details}

\subsection{Language Table Task Details}
\begin{figure}[t]
\centering
\includegraphics[
  width=0.85\linewidth,
  trim=0cm 0cm 0cm 0cm,
  clip
]{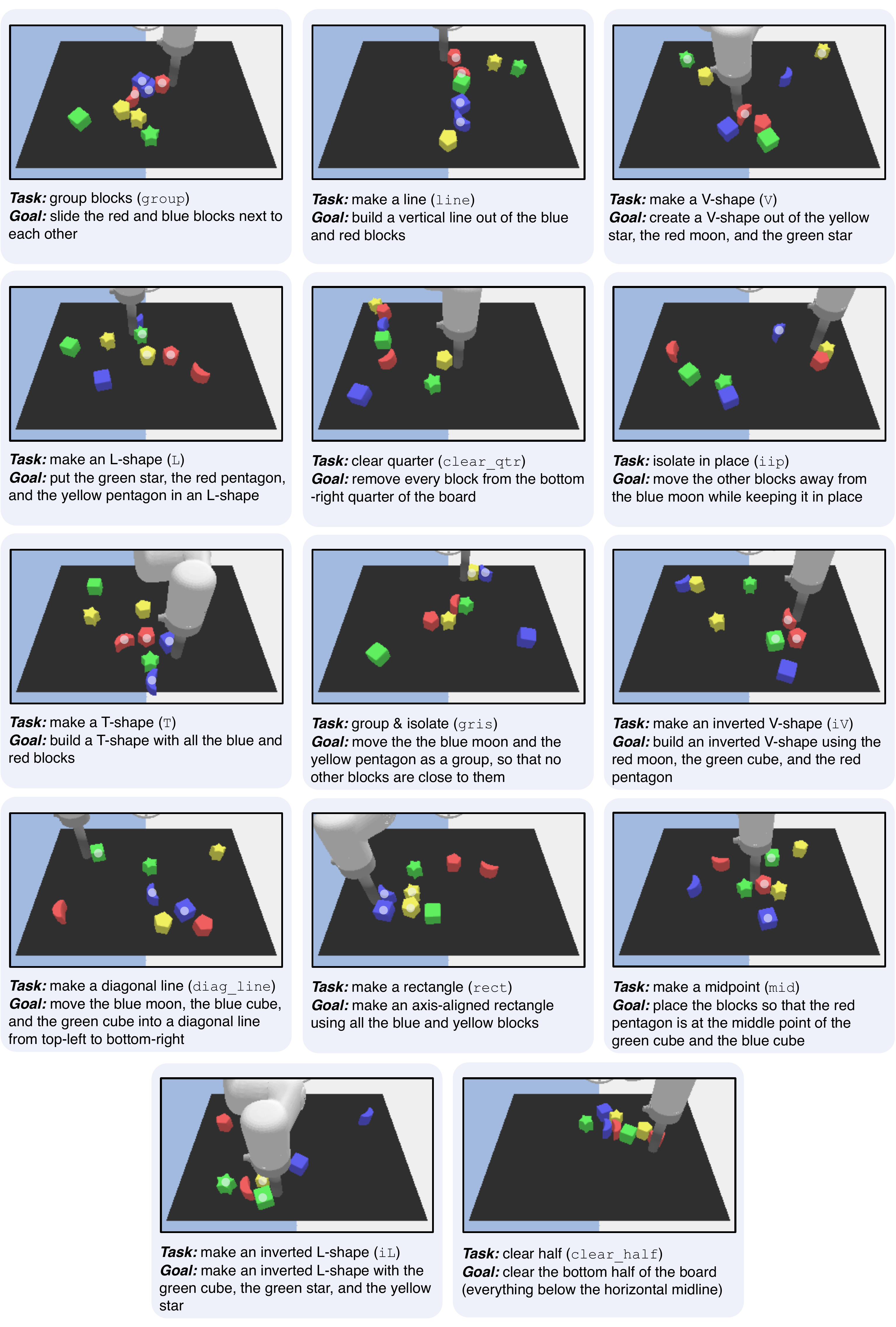}
\caption{\footnotesize{
\textbf{Successful execution of tasks.}
For each task, we show the task name, a long-horizon goal, and the image of the goal state. Task-related blocks are annotated with white dots.
}}
\label{fig:app_14tasks}
\end{figure}

We design 14 long-horizon manipulation tasks in the Language Table environment, which require arranging blocks into task-specific spatial configurations. Each scene contains 8 blocks: \texttt{red moon}, \texttt{red pentagon}, \texttt{blue moon}, \texttt{blue cube}, \texttt{green cube}, \texttt{green star}, \texttt{yellow star}, and \texttt{yellow pentagon}. The robot uses a cylindrical end-effector that pushes blocks on the board. The action space is 2D. 
Figure~\ref{fig:app_14tasks} shows examples of successful execution of tasks for all 14 tasks. We provide full videos of the learned policy at our website \href{https://robotic-reasoner.github.io/}{https://robotic-reasoner.github.io/}.

We next describe the success criteria for each task:

\begin{itemize}[leftmargin=*, itemsep=1pt, topsep=2pt]
  \item \texttt{Group blocks} (\texttt{group}). This task uses 4 blocks. The goal is to move the specified blocks into one compact cluster, so that each block is close to the group's shared center within a certain threshold. 

  \item \texttt{Make a line} (\texttt{line}). This task uses 4 blocks. The goal is to arrange the selected blocks into a straight axis-aligned line, with their perpendicular spread kept within a certain threshold. Depending on the sampled instruction, this line may be horizontal or vertical.

  \item \texttt{Make a V-shape} (\texttt{V}). This task uses 3 blocks. The goal is to arrange the selected blocks into a V shape, where two blocks form a roughly horizontal upper edge and the remaining block sits below their midpoint as the apex. The apex should be centered under the two upper blocks within a certain tolerance, and any selected block can play any role.

  \item \texttt{Make an L-shape} (\texttt{L}). This task uses 3 blocks. The goal is to arrange the selected blocks into an L shape with one corner block, one block extending upward, and one block extending to the right. The two arms should be sufficiently separated from the corner and aligned with the intended vertical and horizontal directions within a certain tolerance.

  \item \texttt{Clear quarter} (\texttt{clear\_qtr}). The goal is to move every block out of the instructed quarter of the board, so that no block center remains inside that region. The target quarter can be any one of the four regions: top-left, top-right, bottom-left, or bottom-right.

  \item \texttt{Isolate in place} (\texttt{iip}). The goal is to keep the target block essentially where it started while moving the surrounding blocks away from it. Success means the target block stays within a small drift threshold of its initial position, and every other block is farther than an isolation threshold from it.

  \item \texttt{Make a T-shape} (\texttt{T}). This task uses 4 blocks. The goal is to arrange the selected blocks into a T shape, with three blocks forming a horizontal top bar and the fourth block forming a vertical stem below the middle of that bar. The stem should be centered under the bar within a certain tolerance, and any selected block may serve as part of the bar or stem.

  \item \texttt{Group \& isolate} (\texttt{gris}). This task uses 2 blocks. The goal is to make the selected blocks into a compact group while keeping other blocks away from that group. Success requires the selected blocks to be close to their shared center within a threshold, and any other block to be farther than an isolation threshold from the selected group.

  \item \texttt{Make an inverted V-shape} (\texttt{iV}). This task uses 3 blocks. The goal is to arrange the selected blocks into an inverted V shape, where two blocks form a roughly horizontal lower edge and the remaining block sits above their midpoint as the apex. The apex should be centered above the two lower blocks within a certain tolerance.

  \item \texttt{Make a diagonal line} (\texttt{diag\_line}). This task uses 3 blocks. The goal is to arrange the selected blocks into a straight diagonal line, with their perpendicular spread kept within a certain threshold. The diagonal variant can run from top-left to bottom-right, or from bottom-left to top-right.

  \item \texttt{Make a rectangle} (\texttt{rect}). This task uses 4 blocks. The goal is to place the selected blocks at the four corners of an axis-aligned rectangle. The rectangle should have a clear horizontal and vertical extent, and each corner should be occupied by one selected block within a certain tolerance.

  \item \texttt{Make a midpoint} (\texttt{mid}). This task uses 3 blocks. The goal is to place the instructed midpoint block at the middle of the segment defined by the two instructed endpoint blocks. The endpoint blocks define the reference line, and the midpoint block should lie near the segment midpoint within a certain threshold relative to the endpoint spacing.

  \item \texttt{Make an inverted L-shape} (\texttt{iL}). This task uses 3 blocks. The goal is to arrange the selected blocks into an inverted L shape with one corner block, one block extending downward, and one block extending to the left. The two arms should be sufficiently separated from the corner and aligned with the intended vertical and horizontal directions within a certain tolerance.

  \item \texttt{Clear half} (\texttt{clear\_half}). The goal is to move every block out of the instructed half of the board, so that no block center remains inside that region. The target half can be the top / bottom half, or the left / right half.
\end{itemize}

\input{appendix/packing}

\vspace{-0.4cm}
\subsection{RL Reward Function}
\label{app:subsec:reward}

The scalar reward $R$ combines an instruction accuracy reward and a length penalty on the response length: $R = R_{\mathrm{acc}} + R_{\mathrm{len}}$.

\textbf{Accuracy reward for Language Table.} A VLM judge compares the parsed model's instruction to the ground truth instruction based on some rubrics, and gives a reward based on whether they match or not. Note that we do not give a separate format reward, and a response will receive a zero accuracy reward if we fail to extract a valid instruction from it. 
\begin{equation}
R_{\mathrm{acc}} =
    \begin{cases}
    1.0  & \text{if exact linguistic match;} \\
    0.5  & \text{if linguistic match with adverb mismatch;} \\
    0.25 & \text{if semantic match;} \\
    0.0  & \text{otherwise}.
    \end{cases}
\end{equation}
If two instructions match in type and all mentioned components, i.e., block names, directions, and adverbs (e.g., ``slightly'' and ``a bit''), they are considered a linguistic match. For example, ``move the blue cube right'' and ``push the blue cube to the right'' form a linguistic match; ``move the blue cube right'' and ``push the blue cube slightly right'' form a linguistic match with an adverb mismatch. If two instructions do not form a linguistic match, but the judge thinks they will lead to the same outcome based on the current scene, then they are considered a semantic match. For example, if the red cube is to the left of the blue moon, then ``push the blue moon into the red cube'' and ``move the blue cube left'' form a semantic match.
Detailed rubrics for linguistic and semantic matches are in Appendix~\ref{app:subsec:vlm_as_judge_prompt}.
We validate this judge against human labels and alternative VLM judges in Appendix~\ref{app:judge_validation}.

\textbf{Accuracy reward for grocery packing.} The parsed model's instruction is given a reward of 1.0 if it matches the ground truth instruction exactly, and 0.0 if not.
\begin{equation}
R_{\mathrm{acc}} =
    \begin{cases}
    1.0  & \text{if exact string match;} \\
    0.0  & \text{otherwise}.
    \end{cases}
\end{equation}

\textbf{Length penalty.} Separately, $R_{\mathrm{len}} \le 0$ is a log-scaled negative term that discourages responses shorter than $T$ words ($n$ = response word count; default $T{=}80$). No penalty is applied once $n \ge T$.

\begin{equation}
R_{\mathrm{len}} =
    \mathrm{clip}\!\left(
        \log_2\!\frac{\mathrm{clip}(n,\,1,\,T)}{T},\;
        -1,\; 0
    \right)
\end{equation}

\vspace{-0.2cm}
\subsection{RL Reward Curve}
Figure~\ref{fig:rl_reward} shows the training reward with a 21-step centered moving average for \ours, \oursQuarterMid, and \oursRLOnly, initialized from full mid-training, 1/4th of the mid-training data, and the base model, respectively. 
The training reward tends to increase with the amount of mid-training performed before RL.

\subsection{Training Hyperparameters and Checkpoint Selection}
We use the LLaMA-Factory framework for SFT~\citep{zheng2024llamafactory}, and the verl framework for RL~\citep{sheng2024hybridflow}. We use Qwen3.5-4B as the base model. Table~\ref{tab:sft_hyperparams} and Table~\ref{tab:rl_hyperparams} show the hyperparameters for mid-training and RL, respectively.
For mid-training in our method, we run 2 epochs of SFT and use the last checkpoint. For the \IL~baselines, to ensure that our comparison is against a strong baseline, we run 4 epochs of SFT for Language Table and 8 epochs of SFT for grocery packing, and evaluate both the last checkpoint and the best checkpoint, i.e., the checkpoint with the lowest validation loss. 
We find that the last checkpoint performs comparably with the best on Language Table, and worse than the best on grocery packing. Therefore, we report the performance of the best checkpoint. 
For RL, we evaluate the checkpoint with the highest mean reward on the validation set for each run.
Figures~\ref{fig:il_checkpoint} and~\ref{fig:rl_checkpoint} visualize checkpoint selection for Language Table, with stars marking the lowest \IL~validation loss and highest \methodname~validation mean reward, respectively.

\vspace{-0.2cm}
\begingroup
\noindent
\centering
\begin{minipage}[t]{0.32\linewidth}
\centering
\includegraphics[width=\linewidth,trim=0 2.0cm 0 1.2cm,clip]{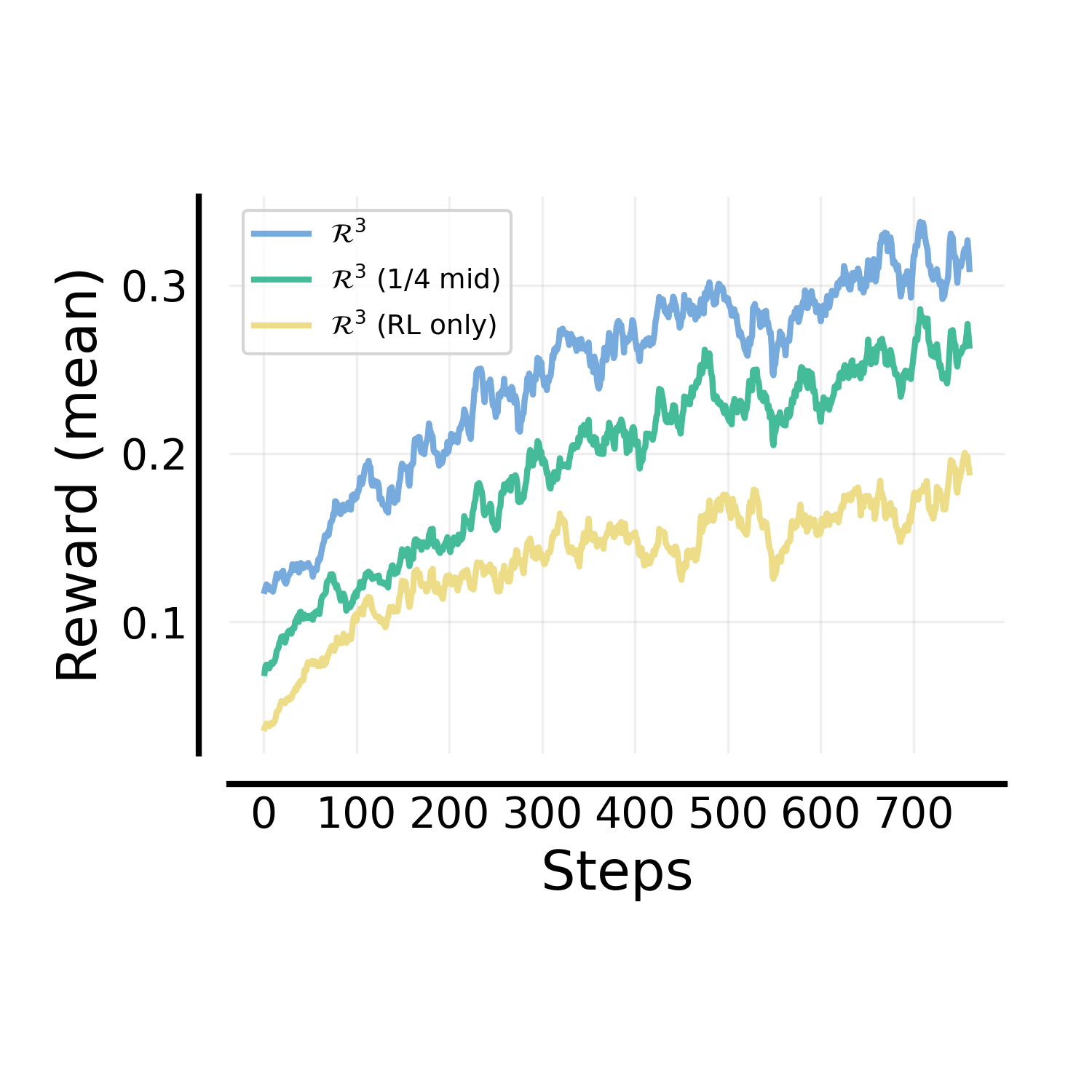}
\captionof{figure}{\footnotesize \methodname~training reward curves for Language Table.}
\label{fig:rl_reward}
\end{minipage}
\hfill
\begin{minipage}[t]{0.32\linewidth}
\centering
\includegraphics[width=\linewidth,trim=0 2.0cm 0 1.2cm,clip]{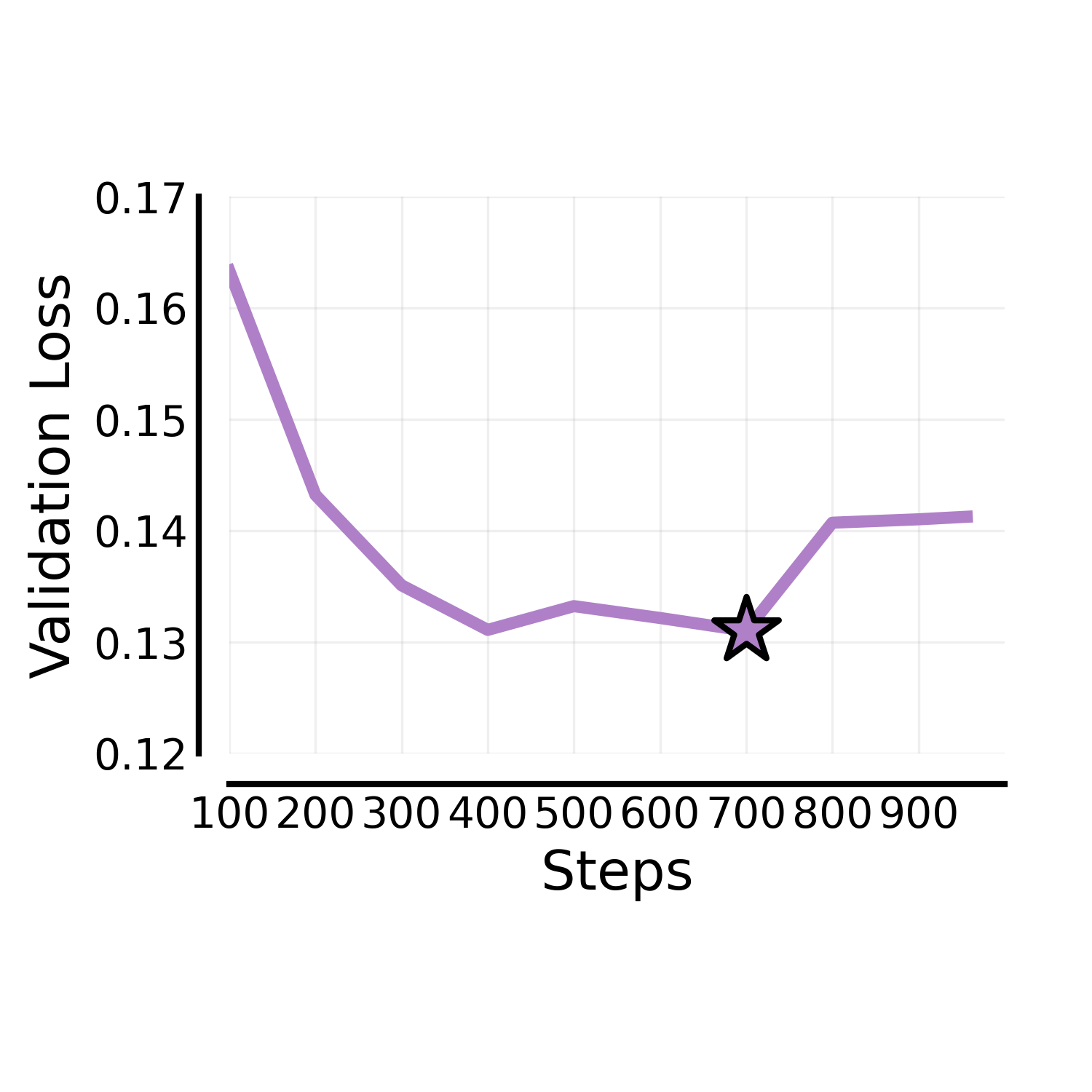}
\captionof{figure}{\footnotesize \IL~validation loss and selected checkpoint for Language Table.}
\label{fig:il_checkpoint}
\end{minipage}
\hfill
\begin{minipage}[t]{0.32\linewidth}
\centering
\includegraphics[width=\linewidth,trim=0 2.0cm 0 1.2cm,clip]{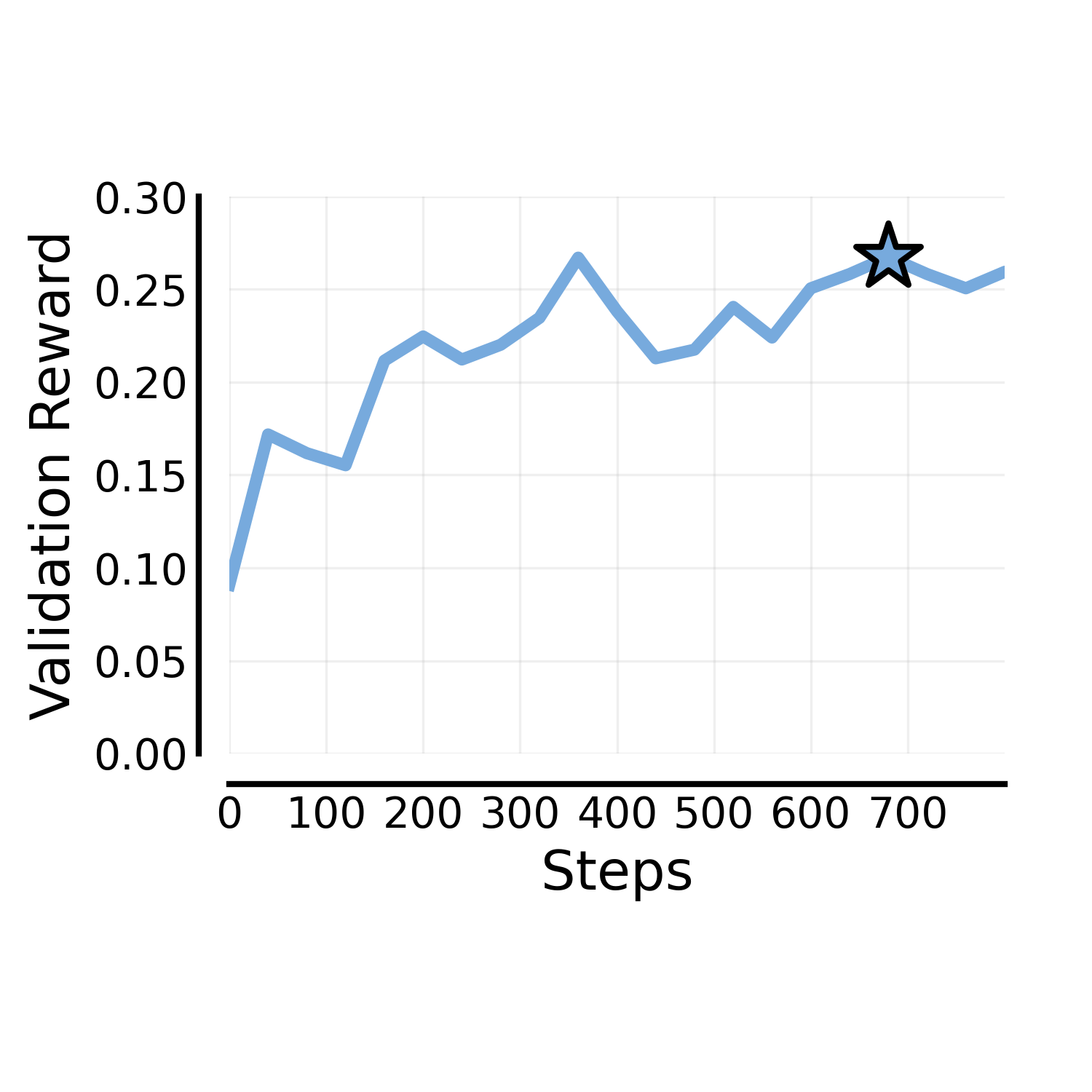}
\captionof{figure}{\footnotesize \methodname~validation reward and selected checkpoint for Language Table.}
\label{fig:rl_checkpoint}
\end{minipage}
\par
\endgroup

\begingroup
\centering
\small
\noindent
\begin{minipage}[t]{0.48\linewidth}
\centering
\begin{tabular}{@{}p{0.4\linewidth}p{0.5\linewidth}}
\toprule
Hyperparameter & Values \\
\midrule
learning rate & $1.0 \times 10^{-6}$ \\
num. train epochs & 2 (mid-train) / \\
& 4 (\IL~for Language Table) / \\
& 8 (\IL~for grocery packing) \\
global batch size & 128 \\
lr scheduler type & cosine \\
warmup ratio & 0.1 \\
finetuning type & full \\
precision & bf16 \\
num.\ GPUs & 8 \\
\bottomrule
\end{tabular}
\vspace{0.1cm}
\captionof{table}{\footnotesize Hyperparameters for SFT.}
\label{tab:sft_hyperparams}
\end{minipage}
\hfill
\begin{minipage}[t]{0.48\linewidth}
\centering
\begin{tabular}{@{}p{0.5\linewidth}p{0.4\linewidth}}
\toprule
Hyperparameter & Values \\
\midrule
learning rate & $2.0 \times 10^{-6}$ \\
num. train epochs & 4 (Language Table) / 8 (grocery packing) \\
train batch size & 32 \\
max response length & 1024 \\
rollouts per prompt & 12 \\
sampling temperature & 1.0 \\
clip ratio (low / high) & 0.2 / 0.3 \\
kl coefficient & 0.0 \\
entropy coefficient & 0.0 \\
\bottomrule
\end{tabular}
\vspace{0.1cm}
\captionof{table}{\footnotesize Hyperparameters for RL.}
\label{tab:rl_hyperparams}
\end{minipage}

\par
\endgroup

\clearpage

\subsection{Our Embodied Chain-of-Thought (ECoT) Implementation}
\label{app:subsec:ecot}

Our goal is to compare with the most faithful approximation of this ECoT approach. Note that our setting and tasks are different from~\citep{zawalski2025robotic} in three important ways: 
\begin{itemize}
    \item The output in our setting is the short-horizon instruction, while the output in~\citep{zawalski2025robotic} is the low-level action. Therefore, in our implementation, we remove the ``move command" part and put the end-effector states and object states before the textual reasoning.
    \item We highlight that our long-horizon manipulation tasks cannot be solved by fixed instruction sequences or planned in an open-loop way, because \textbf{(1)} the goal instance and initial configurations vary across scenes, \textbf{(2)} the low-level policy often fails to follow the instructions and therefore requires closed-loop correction or replanning,  and \textbf{(3)} collisions introduce additional stochasticity. 
    Consequently, supervising the model with explicit future plans would introduce substantial noise, so we omit the ``plan'' component from our implementation. Note that the model can still do planning in textual reasoning.
    \item For reasoning supervision, we record the data collector's reasoning, while ECoT uses post-hoc generated reasoning by querying Gemini for a retrospective rationale that explains the expert instruction. In our implementation, we compare both types of reasoning, so that we disentangle the effect of this component from other ECoT components.
\end{itemize}
Therefore, our implementation of ECoT contains task goal, end-effector state, object states, textual reasoning, and instruction. 
For end-effector and object states, we use their 2D coordinates provided by the Language Table simulation environment.
For textual reasoning, we compare data collector's vs. post-hoc reasoning traces. Data collector's reasoning traces mean the expert reasoning traces used in mid-training of our approach. We follow ECoT to generate post-hoc reasoning traces by asking Gemini 3 Flash to provide a post hoc explanation for the selected instruction. The prompt template is provided in Appendix~\ref{app:subsec:ecot_post_hoc_gen_prompt}. Note that the Gemini model does not have access to its decision-time reasoning when generating post-hoc reasoning. 

\begin{promptbox}{ECoT Response Example}
\small
\begingroup
Goal: Move all blocks out of the top-left quarter of the board.\\
Arm: [0.46, 0.10]\\
Objects:\\
- red moon: [0.23, -0.16]\\
- red pentagon: [0.31, 0.07]\\
- blue moon: [0.30, 0.17]\\
- ... \\
Reasoning: In the current scene, the red moon and blue cube are positioned within the top-left quarter of the board ... \\
Instruction: move the red moon to the center of the board.
\endgroup
\end{promptbox}

\clearpage

%% file: appendix/packing.tex
\subsection{Grocery Packing Task Details}
\label{app:packing}

\parheading{Environment.}
The grocery-packing task suite \citep{anonymous2026vlaexplore} is built in the MuJoCo simulator~\citep{todorov2012mujoco} using two 7-DoF UFACTORY xArm-7 arms with modified gripper fingers~\citep{liu2024fastumi} to pack YCB grocery objects~\citep{calli2015benchmarking} (cracker box, sugar box, tomato soup can, gelatin box, foam brick, and tuna fish can) into small, medium, and large trays.
High-level instructions are pack, remove, or transfer commands over named items and tray sizes.

\parheading{Held-out goals for evaluation.}
\label{app:packing_goals}
Table~\ref{tab:spb_goals} gives the exact high-level goal for each held-out packing specification. We provide full videos of the learned policy at our website \href{https://robotic-reasoner.github.io/}{https://robotic-reasoner.github.io/}.

\begin{table}[H]
\centering
\footnotesize
\setlength{\tabcolsep}{3pt}
\begin{tabular}{@{}lcp{0.9\linewidth}@{}}
\toprule
\textbf{ID} & \textbf{Stages} & \textbf{Goal} \\
\midrule
\taskspack{1} & 2 & Move the gelatin box from the medium tray to the small tray, then pack the foam brick into the small tray. \\
\taskspack{2} & 3 & Pack the gelatin box and foam brick into the medium tray alongside the sugar box. \\
\taskspack{3} & 3 & Remove the cracker box from the medium tray, then pack the sugar box and foam brick there; the gelatin box stays in the small tray. \\
\taskspack{4} & 4 & Move the gelatin box and foam brick from the medium tray to the small tray, then pack the sugar box into the medium tray. \\
\taskspack{5} & 5 & Redistribute five objects from the large tray: cracker box stays in large; sugar box and soup can to medium; gelatin box and foam brick to small. \\
\taskspack{6} & 4 & Move the soup can from small to medium, move the gelatin box and foam brick from medium to small, and pack the sugar box into medium. \\
\taskspack{7} & 2 & Remove the soup can from the small tray so the gelatin box and foam brick fit. \\
\taskspack{8} & 5 & Reorganize: gelatin box and foam brick to small, sugar box and soup can to medium, cracker box to large. \\
\taskspack{9} & 4 & Pack gelatin box and foam brick into medium; sugar box and tuna fish can into large. \\
\taskspack{10} & 4 & Pack soup can into small (upright), sugar box into medium, cracker box and gelatin box into large. \\
\taskspack{11} & 5 & Pack foam brick and sugar box into medium; soup can, tuna fish can, and gelatin box into large. \\
\taskspack{12} & 5 & Pack gelatin box into small (upright), foam brick and sugar box into medium, soup can and cracker box into large. \\
\bottomrule
\end{tabular}
\caption{\footnotesize Goal descriptions and stage counts for 12 held-out packing specifications for evaluation.}
\label{tab:spb_goals}
\end{table}

\parheading{Success criteria.} A goal object counts as packed only if its center of mass lies in its assigned tray and remains below a velocity threshold of \(0.02\)\,m/s for 30 consecutive control steps. Objects in the wrong tray do not count. Designated clutter must occupy no tray at the end of the episode. An orientation constraint requires the object to remain within \(15^\circ\) of the specified upright orientation.
The progress metric is defined as \(K/N\), where \(K\) is the maximum number of stages, i.e., number of correctly packed goal objects, completed during an episode and \(N\) is the number of goal objects.

\parheading{Reasoner training data construction.}
For each segment in human-collected demonstrations, we sample synchronized base and dual-wrist images. We oversample the onset and pre-completion of the subtask and exclude the ambiguous transition tail:
\begin{itemize}[leftmargin=*, itemsep=1pt, topsep=2pt]
    \item starting stage \([0, 1.5\,\mathrm{s})\): 3 frames, including frame zero, with minimum spacing \(0.25\,\mathrm{s}\);
    \item interior stage \([1.5\,\mathrm{s},\, T-2.0\,\mathrm{s})\): up to 4 frames, with minimum spacing \(1.0\,\mathrm{s}\) from all selected frames;
    \item pre-end stage \([T-2.0\,\mathrm{s},\, T-1.0\,\mathrm{s})\): 1 frame, with minimum spacing \(0.25\,\mathrm{s}\);
    \item transition tail \([T-1.0\,\mathrm{s},\, T)\): skipped.
\end{itemize}
Short episodes omit unavailable stages.

\subsection{Comparison of Language Table and Grocery Packing}
\label{app:lt_vs_packing}

Table~\ref{tab:lt_vs_packing} summarizes the domain-specific components of the shared hierarchical reasoning framework.

\begin{table}[H]
\centering
\small
\setlength{\tabcolsep}{4pt}
\begin{tabular}{@{}p{0.27\linewidth}p{0.36\linewidth}p{0.36\linewidth}@{}}
\toprule
 & \textbf{Language Table} & \textbf{Bimanual packing} \\
\midrule
Embodiment & single cylindrical pusher & dual 7-DoF xArm-7 + Robotiq 2F-85 \\
Action & 2D tabletop pushing & 14-D end-effector (7 per arm) \\
Cameras & top-down RGB & base + left/right wrist RGB \\
Low-level policy & pre-trained language-conditioned policy & VLA fine-tuned from \(\pi_{0.5}\) \\
Control frequency & 10\,Hz & 60\,Hz \\
Max episode length & 400 steps (40s) & 21{,}600 steps (300s) \\
Instruction frequency & every 20 steps & every 300 steps \\
Instruction horizon & 400/20=20 & 21{,}600/300=72 \\
RL reward & VLM-as-a-judge & string match \\
\bottomrule
\end{tabular}
\caption{\footnotesize Comparison of key environment and evaluation configs for Language Table and simulated bimanual grocery packing.}
\label{tab:lt_vs_packing}
\end{table}

%% file: appendix/additional_exp.tex
\section{Additional Experimental Results}
\label{app:add_exp_results}

\subsection{Visual Question Answering Evaluation.}
\label{app:subsec:vqa_results}

\parheading{Visual question answering task design.} We first introduce the visual question answering (VQA) task we design to evaluate the perceptual and action-oriented reasoning abilities of different models. We construct 5 classes of VQA questions on Language Table:

\begin{itemize}
  \item \texttt{Absolute Position}. This class asks where a queried block or robot arm is located on the board. The answer is selected from a fixed set of board regions, such as center, top, bottom, left, right, and the four corner regions. The model directly outputs the corresponding region phrase. For example, an input question is ``Where is the red cube located on the board?'' and the ground-truth answer is ``The red cube is in the top left of the board.''

  \item \texttt{Relative Position}. This class asks for the direction of one object relative to another object or the robot arm. The answer is selected from a fixed set of relative directions, including left, right, top, bottom, and the four diagonal directions. The model directly outputs the relative direction phrase. For example, an input question is ``Where is the blue moon relative to the yellow star?'' and the ground-truth answer is ``The blue moon is to the right of the yellow star.''

  \item \texttt{Distance}. This class asks which block is nearest to or farthest from a specified anchor object, where the anchor may be another block or the robot arm. The answer is chosen from the visible block names. The question is not multiple choice; the model outputs the selected block name. For example, an input question is ``Which block is nearest to the arm?'' and the ground-truth answer is ``The green star.''

  \item \texttt{Instruction Execution}. This class asks whether a shown scene correctly satisfies a given instruction. The answer format is binary: correct execution or incorrect execution. This class requires the model to understand the instruction, identify the relevant objects and goal condition, and verify the final visual state. For example, an input prompt is ``... Instruction: move the red cube to the center of the board. Is the instruction correctly and fully executed? A. Correct execution. B. Incorrect execution.'' The response is formatted as ``Think: ... Answer: A'' and evaluation is performed only on the final answer.

  \item \texttt{Instruction Inference}. This class asks which candidate instruction best explains an observed before-and-after visual transition. The answer format is multiple choice over candidate instructions, with the model selecting one option letter. This class tests inverse instruction understanding: the model must infer the intended command from the visual change while distinguishing among similar language choices. For example, an input prompt is ``... Given that the execution is correct, which instruction was executed? A. move the red cube left. B. move the blue moon to the green star. C. separate the yellow pentagon from the blue cube. D... E...'' The response is formatted as ``Think: ... Answer: B'' and evaluation is performed only on the final answer.
\end{itemize}

\begin{table}[h]
\centering
\small
\setlength{\tabcolsep}{5.0pt}
\renewcommand{\arraystretch}{1.05}

\resizebox{\linewidth}{!}{%
\begin{tabular}{@{}l l c @{\hspace{0.8em}} c c c c c c@{}}
\toprule
\textbf{Question Class} &
\textbf{Question Type} &
\textbf{Think?} &
\multicolumn{4}{c}{\textbf{Qwen3.5-4B variants}} &
\multicolumn{2}{c}{\textbf{Reference}} \\
\cmidrule(lr){4-7}\cmidrule(lr){8-9}
& & & Base & \oursMidOnly & \oursRLOnly & \ours & Gemini & Random \\
\midrule

Absolute Position
& Perception & No
& 35.7 & 39.5 & \best{41.8} & 41.3 & 62.3 & -- \\

Relative Position
& Perception & No
& 39.3 & 56.8 & 48.0 & \best{59.3} & 66.7 & -- \\

Distance
& Perception & No
& 29.3 & 48.1 & 38.5 & \best{52.5} & 82.3 & -- \\

Instruction Execution
& Instruction Comp. & Yes
& \best{67.8} & 65.0 & 61.2 & 64.3 & 77.3 & 50.0 \\

Instruction Inference
& Instruction Comp. & Yes
& 30.9 & 45.2 & 43.2 & \best{55.3} & 85.1 & 20.0 \\

\bottomrule
\end{tabular}%
}
\vspace{0.1cm}
\caption{\footnotesize \textbf{Accuracy by VQA question class.} Gemini is reported as an external reference and is not included in the ranking. \textbf{Bold} values mark the best non-reference model.}
\label{app:tab:vqa_results}
\end{table}

\parheading{Visual question answering evaluation results.} Table~\ref{app:tab:vqa_results} shows the performance of different models on each VQA task class. 
The first three classes focus on perception from a static scene: localizing objects on the board, reasoning about pairwise spatial relations, and comparing object distances. The last two classes focus on instruction comprehension from manipulation outcomes: judging whether a visual state satisfies an instruction, and inferring which instruction best explains an observed transition.
Reasoning is required for the Instruction Comprehension tasks, not for the Perception tasks. The analyses of these results are provided in Section~\ref{subsec:representation_learning}.

\subsection{Qualitative Analysis of Reasoning Traces}
\label{app:reasoning_trace_analysis}

To better understand how each training stage changes the model's reasoning behavior, we analyze completions from the base model, the mid-trained model, and the mid-trained + RL model on the same 30 validation scenes. Each completion contains a free-form reasoning trace followed by a high-level instruction. We extract both countable signals, summarized in Table~\ref{tab:reasoning_trace_stats}, and qualitative behavior patterns, summarized in Table~\ref{tab:reasoning_behavior_comparison}.

\begin{table}[h]
\footnotesize
\centering
\setlength{\tabcolsep}{5.0pt}
\renewcommand{\arraystretch}{1.05}

\begin{tabular}{@{}l c c c@{}}
\toprule
\textbf{Signal} & Base & \oursMidOnly & \ours \\
\midrule
Avg. length (chars)
& 1079 & 563 & 696 \\
Max length (chars)
& 4256 & 1288 & 1798 \\
Clean \texttt{Reasoning:}-first format
& 19/30 & 30/30 & 30/30 \\
Backtracking / re-examination
& 5 & 0 & 1 \\
First-person planning
& 11 & 15 & 23 \\
Hallucinated objects
& 3 & 0 & 0 \\
Reference to prior step or plan
& 5 & 2 & 1 \\
\bottomrule
\end{tabular}%
\vspace{0.1cm}
\caption{\footnotesize \textbf{Countable reasoning-trace signals across checkpoints.}
We analyze aligned reasoning traces from the base, mid-trained, and mid-trained + RL models on the same 30 validation scenes. Mid-training stabilizes the output format and removes hallucinated objects, while RL increases explicit planning without reintroducing the base model's formatting failures.}
\label{tab:reasoning_trace_stats}
\end{table}

\begin{table}[p]
\vspace{-0.3cm}
\centering
\small
\setlength{\tabcolsep}{5.0pt}
\renewcommand{\arraystretch}{1.05}

\resizebox{\linewidth}{!}{%
\begin{tabular}{@{}p{0.15\linewidth} p{0.27\linewidth} p{0.25\linewidth} p{0.25\linewidth}@{}}
\toprule
\textbf{Dimension} & Base & \oursMidOnly & \ours \\
\midrule

Output format
& Inconsistent: sometimes places the instruction before reasoning, omits the \texttt{Reasoning:} prefix, includes stray quotes, or fails to emit an instruction.
& Clean and rigid template across all inspected traces.
& Clean and rigid template across all inspected traces. \\

Length / verbosity
& Highly variable, with occasional long rambling traces.
& Shortest and most economical.
& Moderate length: fuller than mid-training, but still controlled. \\

Reasoning style
& Exploratory and deliberative, often thinking through many alternatives.
& Direct and decisive, usually reaching a single-pass conclusion.
& Structured: restates the goal, assesses the current state, then chooses the next step. \\

Self-correction / backtracking
& Frequent second-guessing and occasional non-convergent loops.
& No observed backtracking in the inspected traces.
& Rare backtracking; when present, the model recovers and commits to an instruction. \\

Goal grounding
& Often loses the goal or makes meta-comments about rules and format.
& Briefly restates the task.
& More explicitly re-derives task constraints before acting. \\

Spatial tracking
& Weak; sometimes misreads the scene or confuses object positions.
& Decent; often references the arm position or previous plan.
& Strongest; tracks what has already been placed and what remains to be done. \\

Hallucination
& Sometimes invents objects or claims required objects are missing.
& None observed.
& None observed. \\

Instruction validity
& Several malformed or out-of-spec outputs, including unsupported relations, arm-only moves, or missing instructions.
& Always valid and in-spec in the inspected traces.
& Always valid and in-spec in the inspected traces. \\

Failure mode
& Incoherence or non-termination on harder scenes.
& Occasionally shallow, choosing a plausible move without much verification.
& Occasionally over-reasons before committing. \\

\bottomrule
\end{tabular}%
}
\vspace{0.1cm}
\caption{\footnotesize \textbf{Qualitative reasoning-behavior comparison.}
The base model is exploratory but unstable, mid-training makes the reasoning interface reliable, and RL on top of mid-training produces more deliberate, state-aware reasoning while preserving format and instruction validity.}
\label{tab:reasoning_behavior_comparison}
\vspace{-0.2cm}
\end{table}

Overall, the trajectory is from chaotic-but-creative reasoning in the base model, to terse-and-reliable reasoning after mid-training, to reliable-and-deliberate reasoning after RL. The clearest changes are that mid-training removes most interface-level failures, including malformed formatting and hallucinated objects, while RL increases explicit state-aware planning without reintroducing the base model's instability. 

\clearpage

\subsection{Judge Validation and Sensitivity}
\label{app:judge_validation}

\begin{table}[h]
\centering
\small
\setlength{\tabcolsep}{5.0pt}
\renewcommand{\arraystretch}{1.05}
\begin{tabular}{@{}lccc@{}}
\toprule
& Agree & Cohen's $\kappa$ & Pearson \\
\midrule
Human-maj vs.\ Human$^\dagger$ & 94.7/100 & 0.911 & 0.975 \\
Human-maj vs.\ Qwen3.5-35B-A3B & 90/100 & 0.837 & 0.984 \\
Human-maj vs.\ Gemini 3.6 Flash & 91/100 & 0.842 & 0.974 \\
Human-maj vs.\ GPT-5.6 Sol & 93/100 & 0.880 & 0.985 \\
\bottomrule
\end{tabular}
\vspace{0.1cm}
\caption{\footnotesize Agreement of majority-vote human labels (Human-maj) vs.\ each judge on 100 prompt-response pairs. $^\dagger$Averaged over 3 human annotators.}
\label{app:tab:judge_agreement}
\end{table}

\begin{figure}[h]
\centering
\includegraphics[width=0.28\linewidth]{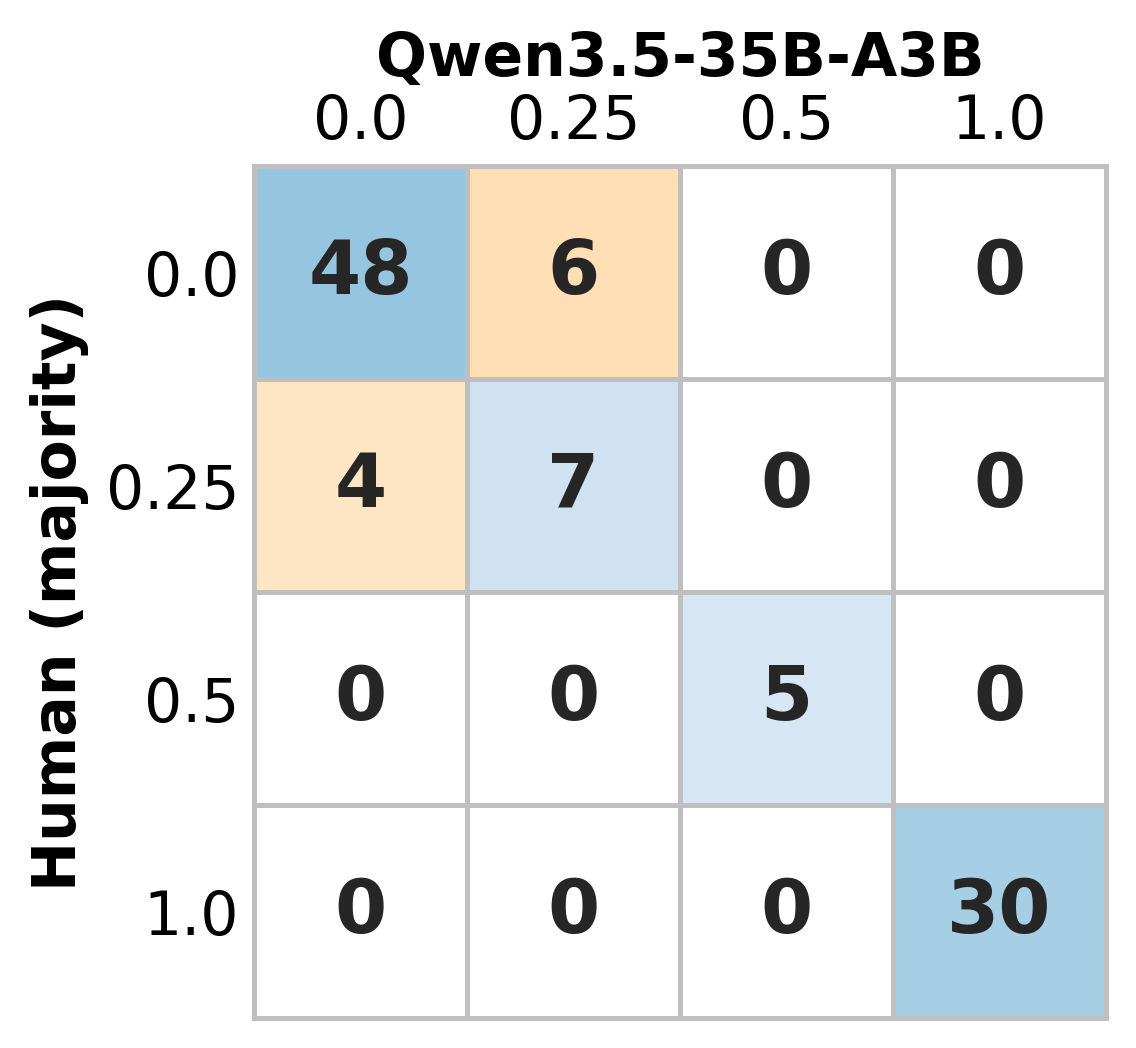}
\caption{\footnotesize Confusion matrix of majority-vote human labels (Human-maj) vs. Qwen3.5-35B-A3B.}
\label{app:fig:judge_confusion}
\end{figure}

We sample 100 prompts from the RL validation set, generate responses with the base model, and score them via 3 human annotators and 3 VLMs, treating the human majority vote as ground truth.
Table~\ref{app:tab:judge_agreement} shows that all VLMs have comparable human agreement, \textit{close to the inter-human reference}, indicating that mismatches largely reflect scene ambiguity rather than judge choice.
Figure~\ref{app:fig:judge_confusion} shows that all mismatches for our Qwen3.5 judge lie between the 0 and 0.25 semantic-match tiers, and these cases are ambiguous even to human annotators. The only clear failure of Qwen3.5 is occasional confusion between ``red pentagon'' and ``red moon.''
Thus, our reward signal is reliable and insensitive to judge choice.

\clearpage

%% file: appendix/examples.tex
\section{Examples}
\label{app:examples}
\subsection{Additional Expert Trajectories}

\begin{figure}[h]
\centering
\includegraphics[width=\linewidth, 
trim=0 0cm 0cm 0,
clip
]{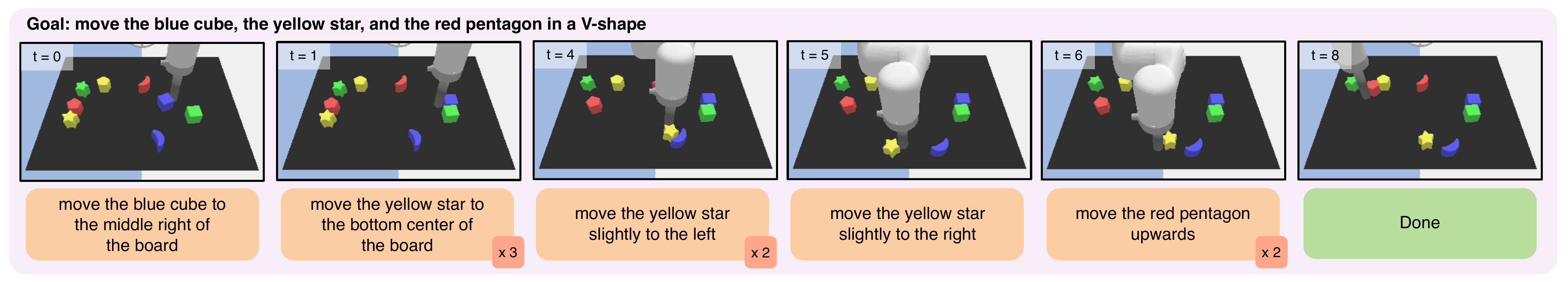}
\caption{\footnotesize{\textbf{Example of expert-collected trajectory on the \taskV~task.} The notation \(\times n\) indicates that the expert repeats this instruction $n$ times.}}
\label{fig:expert_example_app_v}
\end{figure}

\begin{figure}[h]
\centering
\includegraphics[width=\linewidth, 
trim=0 0cm 0cm 0,
clip
]{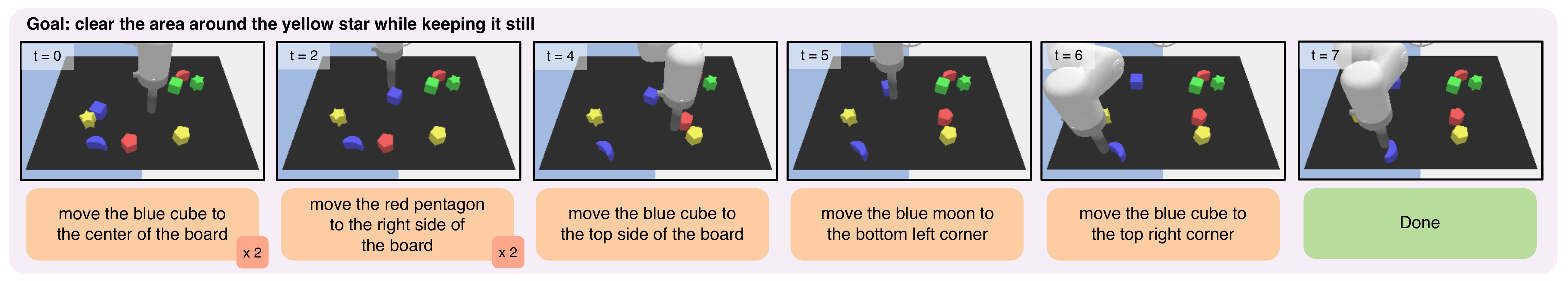}
\caption{\footnotesize{
\textbf{Example expert-collected trajectory on the \taskiip~task.} The notation \(\times n\) indicates that the expert repeats the instruction \(n\) times. At step 6, the instruction is to move the blue cube, but the robot unintentionally pushes the blue moon, resulting in task success. This example shows that expert demonstrations may also contain suboptimal behaviors.
}}
\label{fig:expert_example_app_iip}
\end{figure}

\begin{figure}[h]
\centering
\includegraphics[width=\linewidth, 
trim=0 0cm 0cm 0,
clip
]{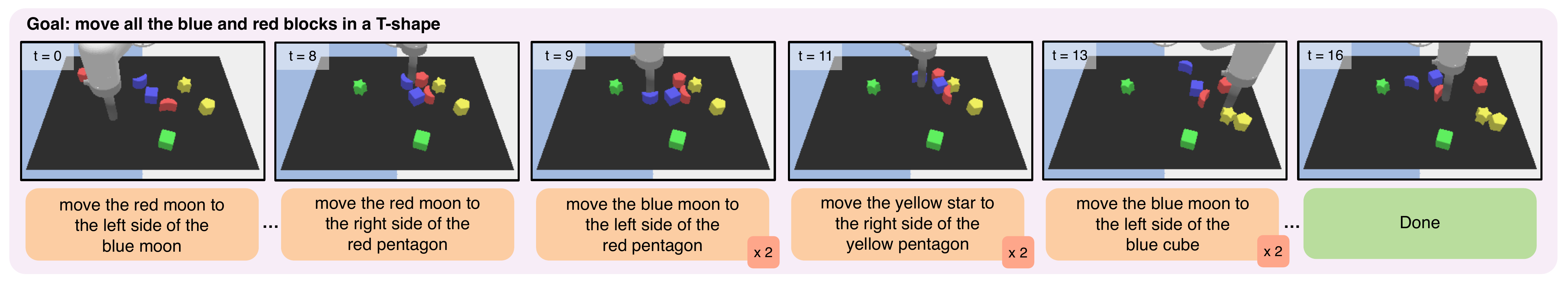}
\caption{\footnotesize{
\textbf{Example expert-collected trajectory on the \taskT~task.} The notation \(\times n\) indicates that the expert repeats the instruction \(n\) times. At step 8, the instruction is to move the red moon, but the robot mistakenly moves the blue moon; therefore, at step 9, the expert first corrects this behavior. At step 11, instead of moving a target block, the expert chooses to clear the yellow star, which is an oracle block for making the T-shape.
}}
\label{fig:expert_example_app_t}
\end{figure}

\clearpage
\vspace{-0.2cm}
\subsection{Additional Reasoning Examples on Language Table}
\vspace{-0.2cm}
\begin{figure}[!htbp]
    \centering
    \includegraphics[width=0.8\linewidth]{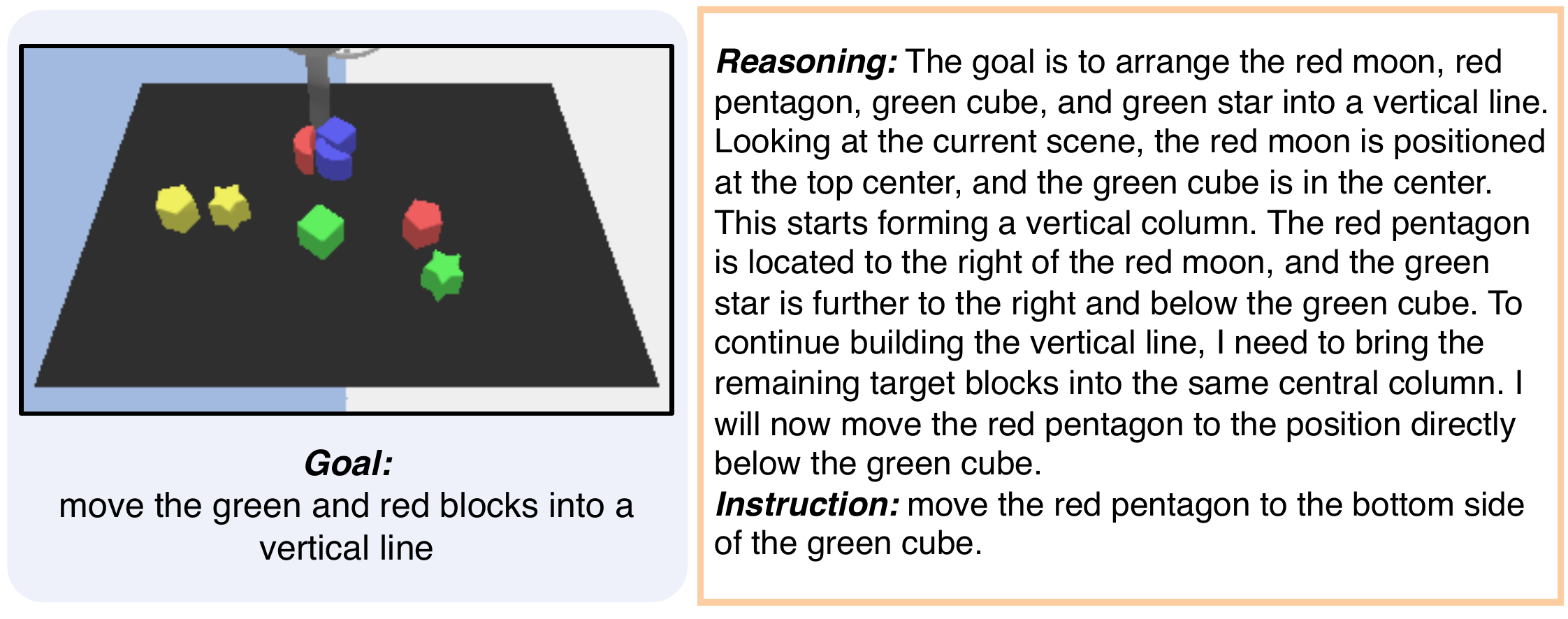}
    \vspace{-0.2cm}
    \caption{\footnotesize \textbf{Example of reasoning strategy learned via \methodname.} 
    The \methodname~model first describes the scene, assesses the current progress toward the goal, plans into the future, and finally outputs an executable instruction.}
    \label{fig:reasoning_example_3_typical}
\end{figure}
\vspace{-0.2cm}
\begin{figure}[!htbp]
    \centering
    \includegraphics[width=0.8\linewidth]{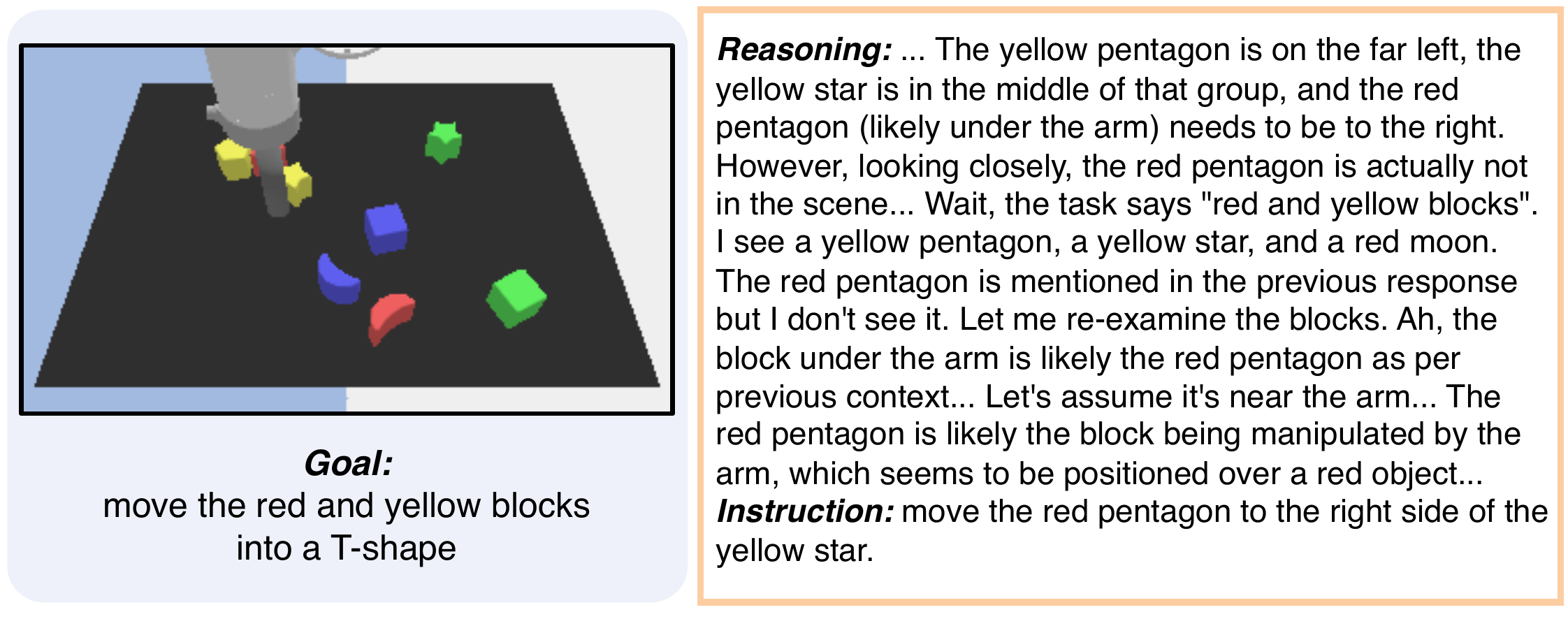}
    \vspace{-0.2cm}
    \caption{\footnotesize \textbf{Example of reasoning strategy learned via \methodname.} 
    The red pentagon is occluded by the robot arm.
    The \methodname{} model re-examines the scene, task information, and history to correctly infer the object state.}
    \label{fig:reasoning_example_2_occlusion}
\end{figure}
\vspace{-0.2cm}
\begin{figure}[!htbp]
    \centering
    \includegraphics[width=0.8\linewidth]{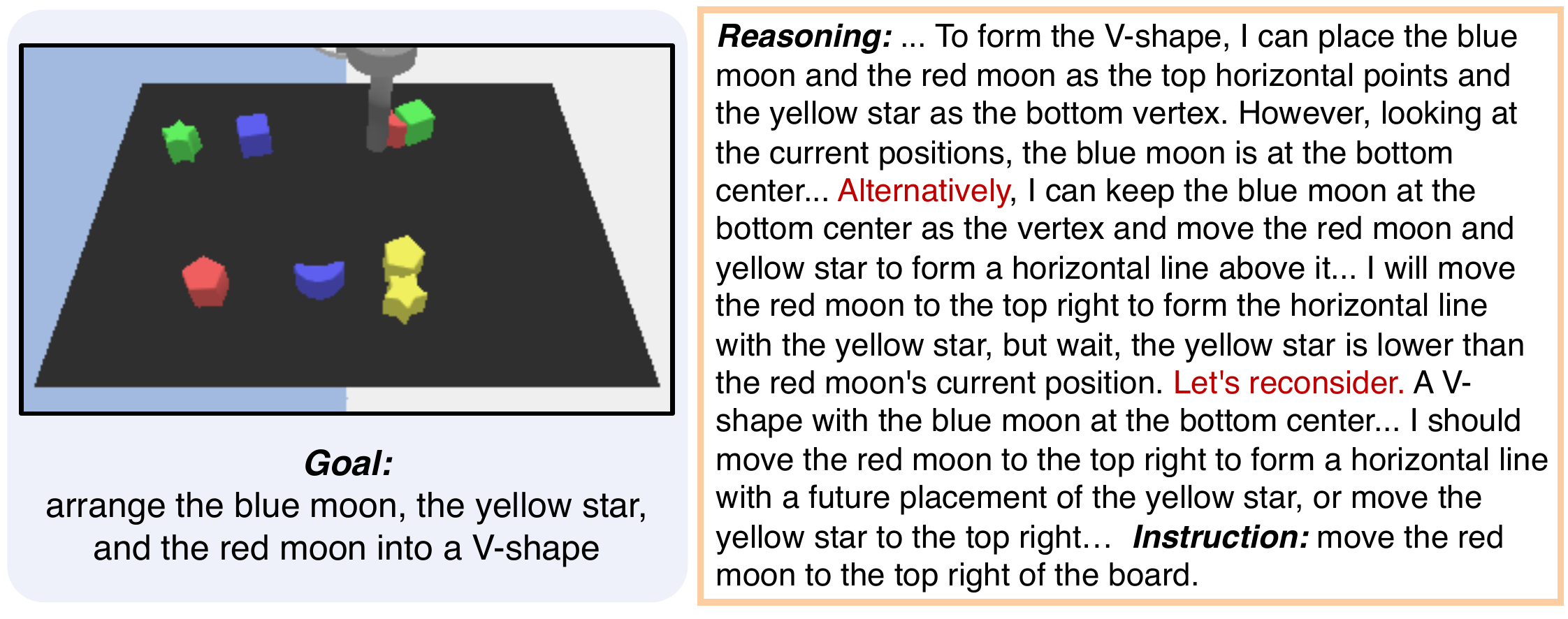}
    \vspace{-0.2cm}
    \caption{\footnotesize \textbf{Example of reasoning strategy learned via \methodname.} The \methodname{} model considers several alternate plans before committing to one particular choice.}
    \label{fig:reasoning_example_1_backtrack}
\end{figure}
\vspace{-0.2cm}
\begin{figure}[!htbp]
    \centering
    \includegraphics[width=0.8\linewidth]{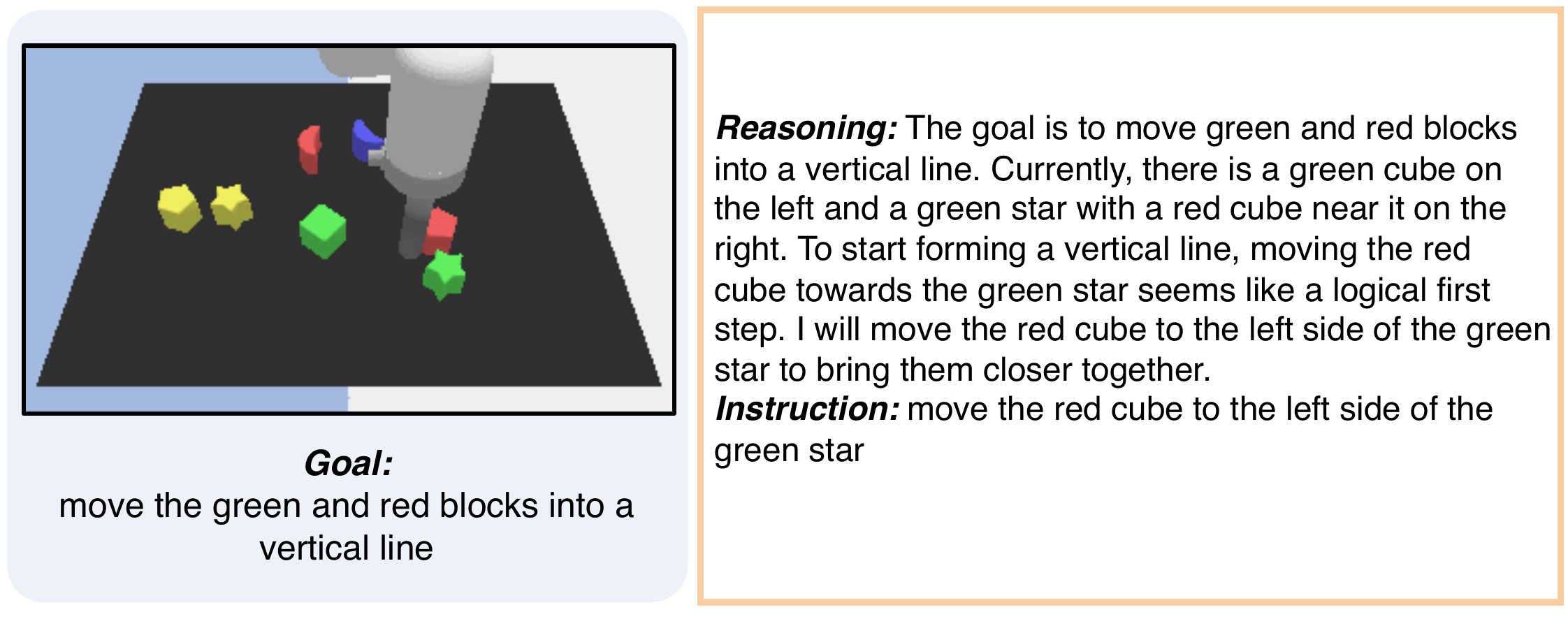}
    \vspace{-0.2cm}
    \caption{\footnotesize \textbf{Example of reasoning strategy of the Qwen3.5-4B base model.} 
    The base model's reasoning shows incomplete scene understanding, weak estimation of task progress, and under-justified planning, leading to a goal-misaligned instruction.}
    \label{fig:reasoning_example_base_typical}
\end{figure}
\vspace{-0.2cm}
\begin{figure}[!htbp]
    \centering
    \includegraphics[width=0.8\linewidth]{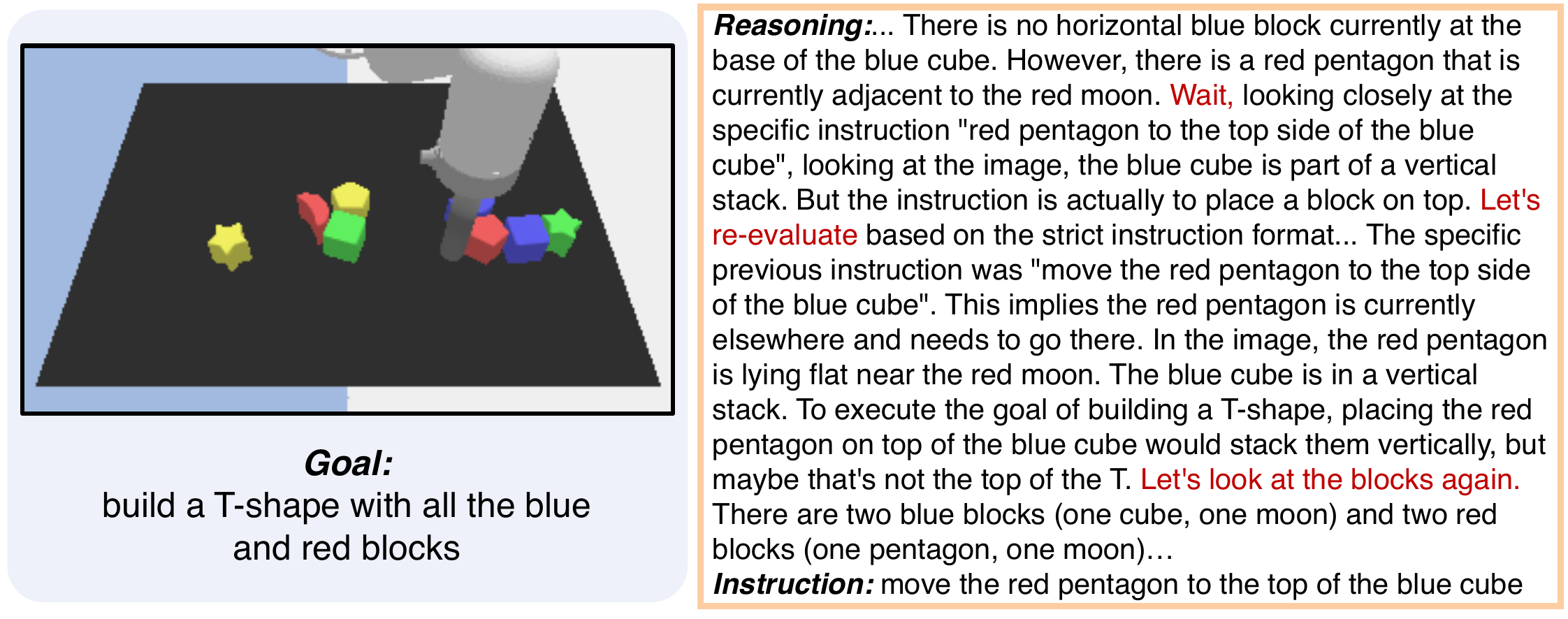}
    \vspace{-0.2cm}
    \caption{\footnotesize \textbf{Example of reasoning strategy of the Qwen3.5-4B base model.} 
    Although the base model sometimes displays backtracking behavior, such behavior is unreliable. The model may signal re-evaluation with phrases such as “Wait” or “Let’s re-evaluate,” but the revision is often superficial or driven by hallucinated assumptions rather than grounded error correction. Consequently, the reasoning becomes inconsistent, poorly structured, and disconnected from the task goal.}
    \label{fig:reasoning_example_base_backtrack}
\end{figure}

\vspace{-0.2cm}
\begin{figure}[!htbp]
    \centering
    \vspace{-0.2cm}
    \includegraphics[width=0.8\linewidth]{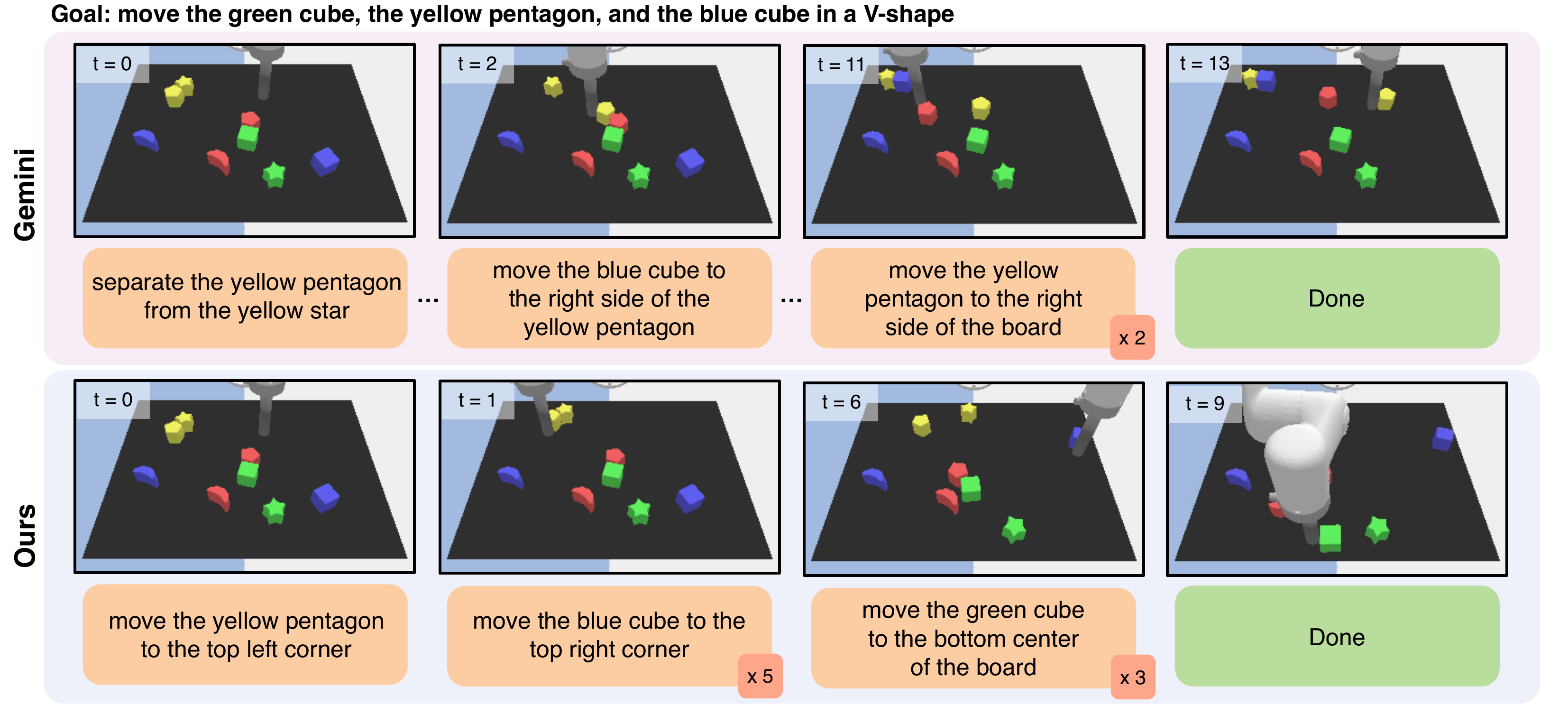}
    \vspace{-0.2cm}
    \caption{\footnotesize{
    \textbf{Comparison of Gemini's and the \methodname{} model's behavior on the \taskV~task.} While Gemini employs more diverse instructions, the \methodname{} model primarily relies on absolute movement instructions.
    }}
    \label{fig:V_traj_comparison}
    \vspace{-0.2cm}
\end{figure}

\clearpage
\subsection{Additional Reasoning Examples on Grocery Packing}
We show qualitative traces from \oursRLOnly{} on held-out packing goals. As on Language Table, the model typically describes the current views, tracks progress toward the long-horizon goal, and then issues a short-horizon instruction; it can also re-examine the scene and revise an initially incorrect interpretation.

\vspace{-0.2cm}
\begin{figure}[!htbp]
    \centering
    \includegraphics[width=0.95\linewidth]{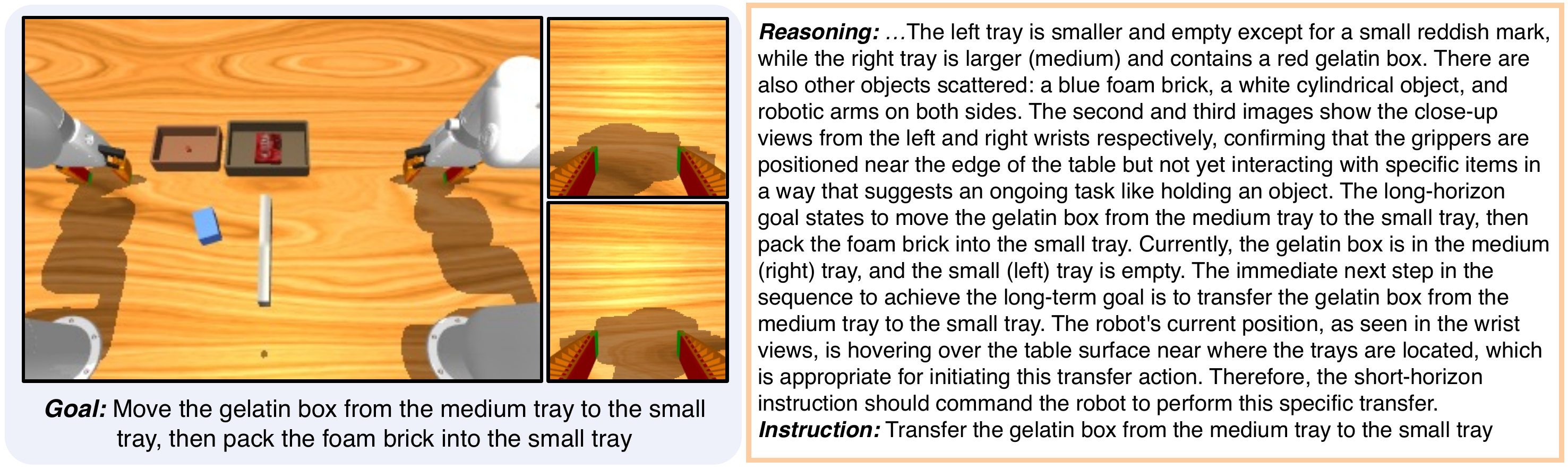}
    \vspace{-0.2cm}
    \caption{\footnotesize \textbf{Example of reasoning strategy learned via \methodname.} 
    The model describes the three camera views, notes that the gelatin box is still in the medium tray, and issues a transfer instruction as the next step.}
    \label{fig:reasoning_example_pack_1_typical}
\end{figure}
\vspace{-0.2cm}
\begin{figure}[!htbp]
    \centering
    \includegraphics[width=0.95\linewidth]{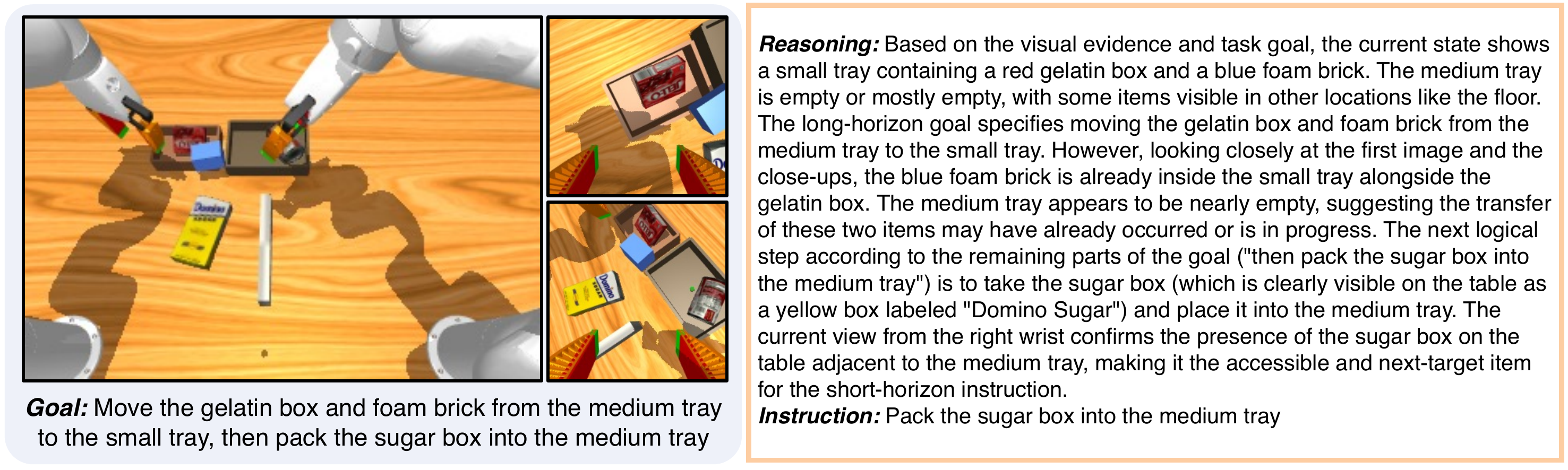}
    \vspace{-0.2cm}
    \caption{\footnotesize \textbf{Example of reasoning strategy learned via \methodname.} 
    The model infers that the gelatin box and foam brick are already in the small tray, so the remaining stage is to pack the sugar box into the medium tray.}
    \label{fig:reasoning_example_pack_2_typical}
\end{figure}
\vspace{-0.2cm}
\begin{figure}[!htbp]
    \centering
    \includegraphics[width=0.95\linewidth]{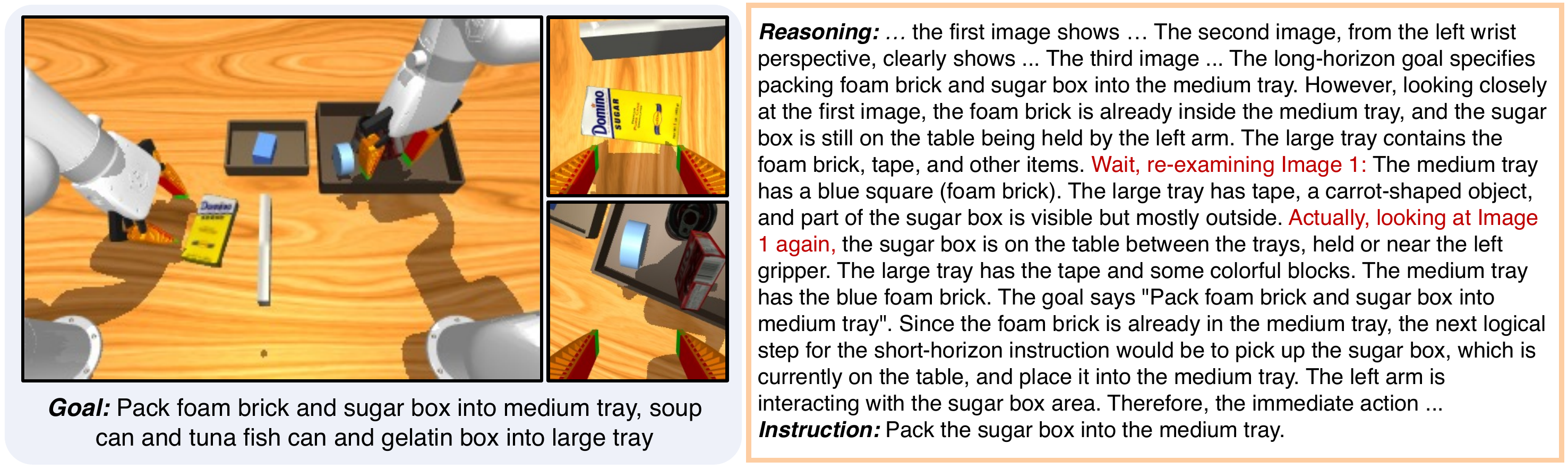}
    \vspace{-0.2cm}
    \caption{\footnotesize \textbf{Example of reasoning strategy learned via \methodname.} 
    Our model first mislocates the foam brick, then re-examines the base view, revises the scene description, and issues the correct next instruction.}
    \label{fig:reasoning_example_pack_3_correction}
\end{figure}

\clearpage

%% file: appendix/prompts.tex
\section{Prompts}
\label{app:prompts}

\subsection{Prompt for the high-level VLM (Language Table)}
The following prompt template is used for Language Table data collection, training, and evaluation.
\begin{promptbox}{Prompt for the high-level VLM (Language Table)}
\small
\begingroup

A robot is performing a long-horizon block arrangement task in a simulator. The robot arm is a gray cylinder that can slide over the board to push colored blocks.

The robot only understands short-horizon instructions.

Your job is to examine the current scene, reason faithfully about what is visible, and then output one short-horizon instruction for the robot.

You are given:
\begin{enumerate}[leftmargin=*, itemsep=0pt, topsep=2pt, parsep=0pt, partopsep=0pt]
    \item The current image of the scene
    \item The long-horizon task goal
    \item Your response at the previous step
\end{enumerate}

\#\#  Task Goal

\{goal\}

Note: \{task\_specific\_prompt\}

\#\#  Your Previous Response

\{previous\_response\}

\#\#  Instruction Guidelines

The robot ONLY understands the following instruction types:
\begin{itemize}[leftmargin=*, itemsep=0pt, topsep=2pt, parsep=0pt, partopsep=0pt]
    \item move $<$block$\_$1$>$ $<$absolute location$>$ \\
    e.g. move the red pentagon to the center of the board / to the top left corner
    \item move $<$block$\_$1$>$ $<$relative direction$>$ of $<$block$\_$2$>$ \\
    e.g. move the yellow star into the top side of the blue moon / move the green star to the left side of the yellow pentagon
    \item push $<$block$\_$1$>$ into $<$block$\_$2$>$ \\
    e.g. push the green star into the yellow pentagon
    \item move $<$block$\_$1$>$ $<$relative direction$>$ \\
    e.g. (slightly) move the red moon upwards / move the blue cube right and down (a bit)
    \item separate $<$block$\_$1$>$ from $<$block$\_$2$>$ \\
    e.g. separate red moon from blue cube
    \item touch $<$block$\_$1$>$ \\
    e.g. touch the green cube
    \item move your arm $<$absolute location$>$ \\
    e.g. move your arm near the bottom center
\end{itemize}

Notes:
\begin{itemize}[leftmargin=*, itemsep=0pt, topsep=2pt, parsep=0pt, partopsep=0pt]
    \item These examples only illustrate valid formats.
    \item You may optionally use adverbs like ``slightly'' or ``a bit'' to control the action magnitude.
    \item Refer to blocks as ``color + shape''. Never say ``the red block'' or ``the cube'' which is ambiguous.
\end{itemize}

\#\#  Reasoning Guidelines

\#\#\# Core Rules
\begin{itemize}[leftmargin=*, itemsep=0pt, topsep=2pt, parsep=0pt, partopsep=0pt]
    \item Use only information visible in the image or explicitly stated.
    \item Do not hallucinate or invent object locations, contacts, completed subgoals, or future outcomes. If the scene is ambiguous or partially occluded, say so and choose the most reasonable action.
    \item Always trust the current observations over the previous response. Your previous response is about the previous step, which might contain hallucination, outdated information, or suboptimal planning. Do NOT blindly follow it.
\end{itemize}

\#\#\# Style
\begin{itemize}[leftmargin=*, itemsep=0pt, topsep=2pt, parsep=0pt, partopsep=0pt]
    \item Provide a concise, explicit stream of consciousness.
    \item Follow strict causality. Avoid presenting the decision before explaining. Instead, your reasoning should naturally yield the instruction.
    \item Avoid unsupported decisions like simply saying ``I want to'' or ``the best instruction is'' without reasoning before it.
\end{itemize}

\#\# Answer Format

``Reasoning: ... Instruction: ...''
\endgroup
\end{promptbox}

\subsection{VLM-as-a-judge Prompt}
\label{app:subsec:vlm_as_judge_prompt}

The following prompt template is used for providing rewards in RL for Language Table. A VLM judge (Qwen3.5-35B-A3B) compares the model's instruction against the ground truth instruction from the expert dataset. The rubrics are elaborated in the prompt. Packing RL does not use this judge; it uses exact instruction-string matching (Appendix~\ref{app:subsec:reward}).

\begin{promptbox}{VLM-as-a-judge Prompt}
\small
\begingroup

A robot performing a block arrangement task on a table. Your task is to compare two textual instructions to the robot and return a score for the candidate.

\#\# Instruction Types

The robot only understands the following types of instructions:
\begin{enumerate}[leftmargin=*, itemsep=0pt, topsep=2pt, parsep=0pt, partopsep=0pt]
    \item \textbf{To absolute location}: move $<$block$\_$1$>$ $<$absolute location$>$. e.g. place the red pentagon at the center of the board / move the green cube to the top left corner
    \item \textbf{To other block's relative direction}: move $<$block$\_$1$>$ $<$relative direction$>$ of $<$block$\_$2$>$. e.g. slide the yellow star into the top side of the blue moon / place the green star to the left side of the yellow pentagon
    \item \textbf{Into other block}: push $<$block$\_$1$>$ into $<$block$\_$2$>$. e.g. push the green star into the yellow pentagon
    \item \textbf{To its own relative direction}: move $<$block$\_$1$>$ $<$relative direction$>$. e.g. (slightly) push the red moon upwards / move the blue cube right and down (a bit)
    \item \textbf{Separation}: separate $<$block$\_$1$>$ from $<$block$\_$2$>$. e.g. separate red moon from the blue cube
    \item \textbf{Touch}: touch $<$block$\_$1$>$. e.g. touch the green cube
    \item \textbf{Arm movement}: move your arm $<$absolute location$>$. e.g. move your arm near the bottom center
\end{enumerate}

\#\# Instructions to Compare

\begin{itemize}[leftmargin=*, itemsep=0pt, topsep=2pt, parsep=0pt, partopsep=0pt]
    \item Candidate instruction: \textbf{\{model\_instruction\}}
    \item Reference instruction: \textbf{\{ground\_truth\_instruction\}}
\end{itemize}

\#\# Evaluation Guidelines

\#\#\# Step 1: Identify the \textbf{linguistic} match.
\begin{itemize}[leftmargin=*, itemsep=0pt, topsep=2pt, parsep=0pt, partopsep=0pt]
    \item Identify the instruction types and components of the candidate and the reference. If \textbf{their types match and all components match}, they form a \textbf{linguistic match}.
    \item Variations allowed:
    \begin{enumerate}
        \item Paraphrasing within ONLY these 4 verbs: "move" = "place" = "slide" = "push".
        \item Paraphrasing of locations or directions with exactly the same meaning, e.g., "top left corner" = "left top of the board", "up" = "upwards", "left" = "to the left" = "to the left side"
        \item Other expression paraphrasing with exact same meanings, e.g., with or without "the", prepositions with same meaning like "into" = "to"
    \end{enumerate}
    \item Variations considered as \textbf{adverb mismatch}: mismatched or missing adverbs like "slightly", "a bit"
    \item Score:
    \begin{itemize}
        \item If they form a linguistic match, skip step 2. Return 1.0 if there is no adverb mismatch, otherwise return 0.5.
        \item If they do not form a linguistic match, go to step 2.
    \end{itemize}
\end{itemize}

\#\#\# Step 2: Identify the \textbf{semantic} match.
\begin{itemize}[leftmargin=*, itemsep=0pt, topsep=2pt, parsep=0pt, partopsep=0pt]
    \item \textbf{Based on the image of the current scene}, imagine the outcome of both instructions.
    \item Two outcomes are considered semantically the same if both conditions are met:
    \begin{enumerate}
        \item they \textbf{move the same block}
        \item the \textbf{final absolute positions} of blocks are the same, though expressed differently.
    \end{enumerate}
    \item Note: two reverse instructions, e.g., "move block\_2 into block\_1" and "move block\_1 into block\_2", are NOT semantically the same because they move different blocks.
    \item Score: If their outcomes are semantically the same, return 0.25. Otherwise, return 0.0.
\end{itemize}

\#\# Output Format

Output a JSON object with the "evaluation" and "score" field.

The evaluation should be a verbose analysis following the guidelines above step by step.

The score should be 1.0 (linguistic match), 0.5 (linguistic match with adverb mismatch), 0.25 (semantic match), or 0.0 (no match).

\endgroup
\end{promptbox}

\subsection{Retrospective Reasoning Generation Prompt}
\label{app:subsec:ecot_post_hoc_gen_prompt}

We query Gemini with the following prompt to generate retrospective reasoning. This is only used in ECoT experiments.

\begin{promptbox}{Retrospective Reasoning Generation Prompt}
\small
\begingroup

A robot is performing a long-horizon block arrangement task in a simulator. The robot arm is a gray cylinder that can slide over the board to push colored blocks.
The robot only understands short-horizon instructions.

I have a dataset of expert demonstrations where the robot follows an expert's short-horizon instructions step by step toward a long-horizon goal. For each step, the expert has already chosen the appropriate instruction. Your job is to write reasoning that explains why that expert demonstration makes sense given the current scene and goal.

\#\# Instruction Types

The robot ONLY understands the following instruction types:
\begin{itemize}[leftmargin=*, itemsep=0pt, topsep=2pt, parsep=0pt, partopsep=0pt]
    \item move $<$block$\_$1$>$ $<$absolute location$>$. \\
    e.g. move the red pentagon to the center of the board / to the top left corner
    \item move $<$block$\_$1$>$ $<$relative direction$>$ of $<$block$\_$2$>$. \\
    e.g. move the yellow star into the top side of the blue moon / move the green star to the left side of the yellow pentagon
    \item push $<$block$\_$1$>$ into $<$block$\_$2$>$. \\
    e.g. push the green star into the yellow pentagon
    \item move $<$block$\_$1$>$ $<$relative direction$>$. \\
    e.g. (slightly) move the red moon upwards / move the blue cube right and down (a bit)
    \item separate $<$block$\_$1$>$ from $<$block$\_$2$>$. \\
    e.g. separate red moon from blue cube
    \item touch $<$block$\_$1$>$. \\
    e.g. touch the green cube
    \item move your arm $<$absolute location$>$. \\
    e.g. move your arm near the bottom center
\end{itemize}

\#\# Guidelines

\begin{itemize}[leftmargin=*, itemsep=0pt, topsep=2pt, parsep=0pt, partopsep=0pt]
    \item The expert's chosen instruction is provided for your reference only. Do not quote or mention it. Write as if you are the expert deciding the next step: reason through the scene and goal step by step, so the appropriate instruction emerges as the conclusion rather than the premise.
    \item The expert's chosen instruction is what the robot should do next, not what it is currently doing.
    \item I provide your previous response to help you better understand the robot's state. Note that the robot may not have successfully executed the previous instruction, so always trust the image over the previous response about the progress.
    \item In your reasoning, be specific about positions, movements, and spatial relations of objects and the arm.
    \item Do not start with 'The goal is...' or similar. The goal is known.
\end{itemize}

\#\# Examples

\begin{enumerate}[leftmargin=*, itemsep=0pt, topsep=2pt, parsep=0pt, partopsep=0pt]
    \item "The blue moon and blue cube are already vertically aligned in the center of the board. The green cube and green star are currently located to the right of the blue blocks. The robot arm's pusher is positioned between the green star and the green cube, ready to interact with the latter. To continue building the vertical line, the green cube should be moved so that it is positioned directly below the blue cube."
    \item "Currently, the red pentagon is on the left side of the board. The blue cube is near the center, horizontally between the red pentagon and the red moon. To form the V, the blue cube needs to be moved to the right of the red moon. Additionally, the blue cube is currently lower (closer to the camera) than the red pentagon, so it must move upwards (further from the camera) to align horizontally with the red pentagon."
\end{enumerate}

\#\# Your Job

Provide a reasoning for the following:
\begin{itemize}[leftmargin=*, itemsep=0pt, topsep=2pt, parsep=0pt, partopsep=0pt]
    \item Task Goal: \{goal\}
    \item Note: \{task\_specific\_prompt\}
    \item The included image shows the robot's current observation.
    \item Your Previous Response: \{previous\_response\}
    \item The expert's chosen instruction: \{instruction\}
\end{itemize}

\endgroup
\end{promptbox}

\subsection{Prompt for the high-level VLM (grocery packing)}
\label{app:packing_prompt}
The following prompt template is used for packing training and evaluation of reasoning models. Instruction-only (no-think) variants omit the reasoning guidelines and ask the model to output only \texttt{Instruction: ...}. Interaction history is the previous instruction rather than the previous full response. Packing RL rewards the parsed instruction by exact string match against the ground-truth instruction (Appendix~\ref{app:subsec:reward}); there is no packing VLM-as-a-judge prompt.

\begin{promptbox}{Prompt for the high-level VLM (grocery packing)}
\small
\begingroup

A bimanual robot is performing a long-horizon packing task.
The robot only understands short-horizon instructions.

Your job is to examine the current images, reason faithfully about what is visible, and then output one short-horizon instruction for the robot.

You are given:
\begin{itemize}[leftmargin=*, itemsep=0pt, topsep=2pt, parsep=0pt, partopsep=0pt]
    \item Three current camera views:
    \begin{itemize}[leftmargin=*, itemsep=0pt, topsep=1pt, parsep=0pt, partopsep=0pt]
        \item The first image is the base camera view
        \item The second image is the left wrist camera view
        \item The third image is the right wrist camera view
    \end{itemize}
    \item The long-horizon task goal
    \item The previous instruction
\end{itemize}

\#\# Task Goal

\{goal\}

\#\# Previous Instruction

\{previous\_instruction\}

\#\# Instruction Guidelines

Choose from one of the following instruction types:
\begin{itemize}[leftmargin=*, itemsep=0pt, topsep=2pt, parsep=0pt, partopsep=0pt]
    \item Pack the $<$item$>$ into the $<$size$>$ tray. Example: Pack the can opener into the medium tray.
    \item Remove the $<$item$>$ from the $<$size$>$ tray. Example: Remove the tuna fish can from the medium tray.
    \item Transfer the $<$item$>$ from the $<$size$>$ tray to the $<$size$>$ tray. Example: Transfer the coffee mug from the medium tray to the large tray.
\end{itemize}

Choose $<$item$>$ from: can opener, candy box, coffee mug, cracker box, foam brick, gelatin box, nesquik canister, soup can, sugar box, tuna fish can

Choose $<$size$>$ from: large, medium, small

\#\# Reasoning Guidelines

\#\#\# Core Rules
\begin{itemize}[leftmargin=*, itemsep=0pt, topsep=2pt, parsep=0pt, partopsep=0pt]
    \item Use only information visible in the images or explicitly stated.
    \item Do not hallucinate or invent object locations, contacts, completed subgoals, or future outcomes. If the scene is ambiguous or partially occluded, say so and choose the most reasonable action.
\end{itemize}

\#\#\# Style
\begin{itemize}[leftmargin=*, itemsep=0pt, topsep=2pt, parsep=0pt, partopsep=0pt]
    \item Provide a concise, explicit stream of consciousness.
    \item Follow strict causality. Avoid presenting the decision before explaining. Instead, your reasoning should naturally yield the instruction.
    \item Avoid unsupported decisions like simply saying ``I want to'' or ``the best instruction is'' without reasoning before it.
\end{itemize}

\#\# Answer format

``Reasoning: ... Instruction: ...''

Do NOT output anything else after the instruction.

\endgroup
\end{promptbox}

\clearpage